%% file: paper.tex
\documentclass[11pt]{article}

\usepackage[final]{acl}

\usepackage{times}
\usepackage{latexsym}

\usepackage[T1]{fontenc}

\usepackage[utf8]{inputenc}

\usepackage{microtype}

\usepackage{inconsolata}

\usepackage{graphicx}
\usepackage{amsmath}
\usepackage{amssymb}
\usepackage{booktabs}
\usepackage{enumitem}
\usepackage{xspace}
\usepackage{multirow}
\usepackage{algpseudocode}
\usepackage[most]{tcolorbox}
\newcommand{\ph}[1]{\textcolor{blue}{\texttt{<#1>}}}
\newcommand{\promptHead}[1]{\par\smallskip\noindent\textbf{#1}\hspace{0.5em}\ignorespaces}
\newtcolorbox{promptbox}{
  enhanced, breakable,
  colback=gray!8, colframe=gray!55,
  boxrule=0.5pt, arc=2pt,
  left=8pt, right=8pt, top=6pt, bottom=6pt,
  before skip=6pt, after skip=6pt,
  fontupper=\small\ttfamily,
  before upper={\let\paragraph\promptHead},
}
\usepackage{dblfloatfix}
\usepackage[linesnumbered,ruled,lined]{algorithm2e}
\usepackage[misc]{ifsym}

\SetCommentSty{mycommfont}

\newcommand{\mat}[1]{{\bf #1}}   

\newcommand{\method}{\textsc{MetaCaster}\xspace}
\newcommand{\firstagent}{\textsc{MGagent}\xspace}
\newcommand{\secondagent}{\textsc{FTagent}\xspace}
\newcommand{\thirdagent}{\textsc{HPagent}\xspace}
\newcommand{\library}{\textsc{LT-Lib}\xspace}

\newcommand{\best}[1]{\textcolor{red}{\textbf{#1}}}
\newcommand{\second}[1]{\textcolor{blue}{\underline{#1}}}

\newcommand\blfootnote[1]{%
  \begingroup
  \renewcommand\thefootnote{}\footnote{#1}%
  \addtocounter{footnote}{-1}%
  \endgroup
}

\title{\method: Meta-Harness-Optimized Agent for End-to-End\\
Few-Shot Learning of Lightweight Time Series Forecasters}

\author{ChengAo Shen\textsuperscript{\rm 1}, Wenchao Yu\textsuperscript{\rm 2}, Fangyu Wu\textsuperscript{\rm 3}, Dongjin Song\textsuperscript{\rm 4}, Hanghang Tong\textsuperscript{\rm 5},\\
\textbf{Dongsheng Luo\textsuperscript{\rm 6}, Wei Cheng\textsuperscript{\rm 2}, Haifeng Chen\textsuperscript{\rm 2}, Jingchao Ni\textsuperscript{\rm 1}}\textsuperscript{\Letter}\\
\textsuperscript{1}University of Houston, \textsuperscript{2}NEC Labs, \textsuperscript{3}University of Waterloo, \textsuperscript{4}University of Connecticut,\\
\textsuperscript{5}University of Illinois at Urbana-Champaign, \textsuperscript{6}Singapore Management University\\
\textsuperscript{\rm 1}\texttt{\{cshen9, jni7\}@uh.edu}, \textsuperscript{\rm 2}\texttt{\{wyu, weicheng, haifeng\}@nec-labs.com},\\
\textsuperscript{\rm 3}\texttt{fangyu.wu@uwaterloo.ca}, \textsuperscript{\rm 4}\texttt{dongjin.song@uconn.edu},\\
\textsuperscript{\rm 5}\texttt{htong@illinois.edu}, \textsuperscript{\rm 6}\texttt{dsluo@smu.edu.sg}}

\begin{document}

\maketitle
\input{Sec/abs}
\input{Sec/intro}

\input{Sec/relate}
\input{Sec/method}

\input{Sec/exp}

\input{Sec/conclusion}

\clearpage
\input{Sec/limitations}
\input{Sec/ack}

\bibliography{ref}

\clearpage
\input{Sec/app}

\end{document}

%% file: Sec/abs.tex
\begin{abstract}
Time series forecasting (TSF) is evolving toward multimodal and agentic settings, yet using foundation models remains uneconomical in resource-constrained scenarios, where compact, specialized forecasters are more desirable. However, lightweight forecasters typically require substantial training data, limiting their use in domains with scarce, slowly accumulated, or privacy-sensitive time series. To address this dilemma, we investigate the challenging problem of few-shot learning for lightweight forecasters. We propose \method, a meta-harness-optimized multi-agent framework that uses agentic data generation to automatically train specialized lightweight forecasters from only a few examples and textual contexts. Our work highlights a new TSF paradigm in which agents act not as forecasters but as intermediary engineers that prepare efficient, task-specific forecasters for deployment. Experiments on 18 datasets, 23 state-of-the-art lightweight forecasters, and 14 baselines demonstrate that \method achieves both data efficiency and computational efficiency while maintaining high-quality TSF performance.\textsuperscript{1}\blfootnote{
\textsuperscript{\Letter}Corresponding author. \textsuperscript{1}The code is available at \url{https://github.com/D2I-Group/metacaster}.}

\end{abstract}

%% file: Sec/intro.tex

\section{Introduction}\label{sec:intro}

\begin{figure*}[t]
  \centering
  \includegraphics[width=0.9\linewidth]{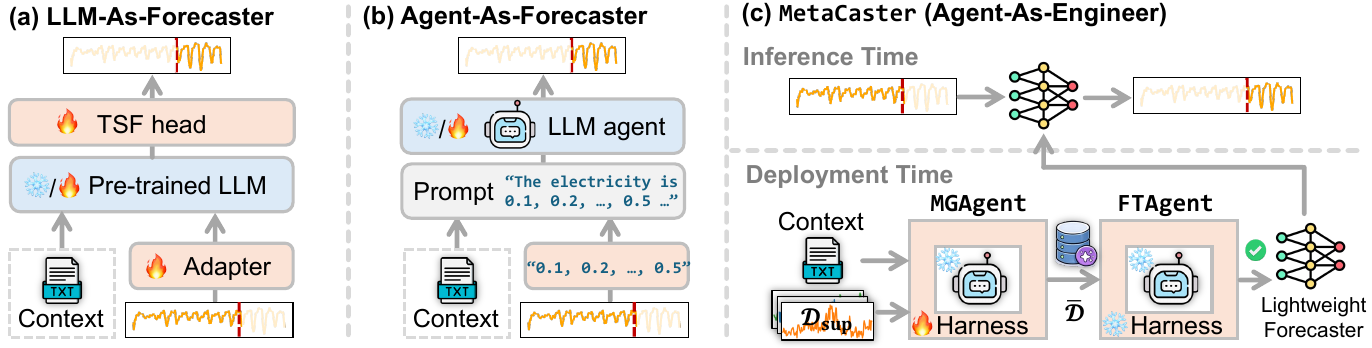}
  \caption{Comparison of different paradigms of using LLMs for TSF. \method is the proposed model.}\label{fig.intro}
  \vspace{-0.1cm}
\end{figure*}

Time series forecasting (TSF) underpins decision-making and intelligence across domains such as geoscience, healthcare, and energy \cite{koprinska2018convolutional,morid2023time,ardid2025ergodic}. Unlike traditional TSF, which focuses on numerical prediction at future timesteps, emergent TSF settings are multimodal, incorporating textual context describing domains, events, or external conditions \cite{jiang2025multi}. This transition is driven by advances in large language models (LLMs) and AI agents. As Fig. \ref{fig.intro} illustrates, recent LLM-based TSF methods have two main paradigms: (a) {\em LLM-as-Forecaster} -- where pre-trained LLMs serve as forecasting backbones with additional cross-modal adapters and TSF heads ({\em e.g.}, \texttt{TimeLLM} \cite{jin2024timellm}, \texttt{S2IPLLM} \cite{pan2024s}, \texttt{TimeVLM} \cite{zhong2025time}, {\em etc.}); and (b) {\em Agent-as-Forecaster} -- where LLM agents interpret time series in contextual prompts, reason about future trends, and generate forecasts conversationally or visually ({\em e.g.}, \texttt{LLMTime} \cite{gruver2023llmtime}, \texttt{TimeOmni-1} \cite{guan2025timeomni}, \texttt{Nexus} \cite{das2026nexus}, {\em etc.}).

Despite the inspiring progress, using LLMs as the computational core for forecasting is largely obstructed by the modality gap between {\em discrete} language tokens and {\em continuous} time series values. Consequently, the effectiveness of the paradigms in Fig. \ref{fig.intro}(a) and (b), compared with specialized non-LLM forecasters, remains under debate \cite{tan2024language,merrill2024language}. Benchmark studies consistently show that dedicated numerical models, such as \texttt{DLinear} \cite{zeng2023dlinear} and \texttt{PatchTST} \cite{nie2023patchtst}, can outperform LLM-based approaches \cite{shen2026multi}, motivating the rise of time series foundation models (TSFMs) such as \texttt{Moirai} \cite{liu2024moirai}, \texttt{Chronos} \cite{ansari2024chronos}, and \texttt{Sundial} \cite{liu2025sundial}.


However, using large models -- whether LLMs or TSFMs -- as one-size-fits-all solutions raises concerns about sustainability and carbon footprint \cite{bolon2024review}. The demand for rapid deployment in resource-constrained settings ({\em e.g.}, edge devices and small organizations) has therefore sparked growing interest in lightweight forecasters \cite{li2024mixlinear}, analogous to the rise of small language models (SLMs) \cite{belcak2025small}. Notably, recent studies show that lightweight forecasters can rival LLMs and TSFMs in individual TSF tasks \cite{shen2025svtime}, their main limitation is the need for sufficient downstream training data, as they lack large-scale pre-training.


In practice, collecting long-history time series for training can significantly delay deployment. In privacy-sensitive domains such as healthcare and finance, acquiring large training datasets may even be infeasible. This raises a challenging question: {\em can lightweight forecasters be effectively trained with only a few samples?} Although such models are highly prone to overfitting without large-scale pre-training, we demonstrate that a solution is possible in the era of agentic AI.


To address this challenge, we propose \method, a \underline{Meta}-harness-optimized agent for end-to-end few-shot learning of lightweight fore\underline{caster}s. As Fig. \ref{fig.intro}(c) illustrates, rather than directly generating forecasts, \method's agents act as intermediaries that prepares a specialized forecaster for a target TSF task. At deployment time, given a few-shot support set $\mathcal{D}_{\text{sup}}$ and contextual description $\mathsf{C}$, \method employs two agents: (1) \firstagent refers to $\mathcal{D}_{\text{sup}}$ and $\mathsf{C}$, and generates a sufficient dataset $\bar{\mathcal{D}}$ that complies with domain constraints; and (2) \secondagent trains and selects the best lightweight forecasters using $\bar{\mathcal{D}}$. To support \secondagent, we compile 23 state-of-the-art (SOTA) lightweight forecasters (2022-2026) into \library library (\S\ref{sec.secondagent}) with a unified API. Similar to fine-tuning a foundation model, \method performs task adaptation at deployment time, but more efficiently via the novel agents and LLM API calls. After deployment, only the selected lightweight forecaster is used for inference.


Unlike existing time series generation models that focus on data simulation \cite{timedp2025,verbalts2025,t2s2025}, \method generates time series specifically to improve forecasting performance. To our knowledge, it is the first framework to align data generation with forecaster quality. Motivated by recent findings on the importance of agent Harnesses \cite{li2026agentharness,metaharness2026} -- the infrastructure surrounding an LLM ({\em e.g.}, system prompts, skills, tools) -- we optimize \firstagent's Harness using a meta-harness \thirdagent (Fig. \ref{fig:method}), which is more efficient than fine-tuning LLMs. The optimized Harness is also transferable across different LLMs, enabling flexible API switching in downstream deployment, as demonstrated in \S\ref{ssec:exp-main}. In summary, our contributions are as follows.
\begin{itemize}[noitemsep,topsep=0pt,leftmargin=*]
\item We investigate the challenging problem of few-shot learning for lightweight forecasters.
\item We propose \method, a novel meta-harness-optimized multi-agent framework that achieves large-model-like performance with efficient inference for TSF tasks.
\item We compile 20+ SOTA lightweight forecasters into \library, a unified library 
released alongside \method.
\item We conduct comprehensive experiments on 18 datasets against 14 baselines, demonstrating the effectiveness of \method.
\end{itemize}

%% file: Sec/relate.tex
\section{Related Work}
\label{sec:relate}

\noindent{\textbf{LLM-based TSF}}. As discussed in \S\ref{sec:intro}, many existing LLM-based TSF models adopt either {\em LLM-As-Forecaster} \cite{zhou2023gpt4ts,jin2024timellm,pan2024s,liu2024autotimes,liu2025calf,chattime2025,zhong2025time} or {\em Agent-As-Forecaster} \cite{xue2023promptcast,gruver2023llmtime,das2026nexus}. Additionally, some multi-task time series QA agents can conduct TSF via LLM reasoning \cite{kong2025time,guan2025timeomni,tsreasoner2025,wu2026timeart,guan2026timeomni}. In contrast, there are relatively fewer {\em Agent-As-Engineer} models that use Harness for TSF \cite{zhao2025timeseriesscientist,timecopilot2025,alphacast2025,flairrts2025}. However, these models don't generate times series (thus require large training data), and never automatically optimize their Harness for TSF, distinguishing them from \method.




\vspace{0.1cm}

\noindent{\textbf{Time Series Generation}}. Our work is related to time series generation models, including data augmentation techniques \cite{luo2023time,yue2022ts2vec,iwana2021empirical,wen2021ts_aug}, generative models that focus on data distributions \cite{yoon2019timegan,desai2021timevae,jeon2022gtgan,yuan2024diffusionts}, and language-based models that can encode contexts \cite{naiman2024imagentime,timedp2025,verbalts2025,t2s2025}. However, these standalone generators aim to simulate certain data properties, rather than directly optimize TSF performance, leaving a significant gap as we will demonstrate in \S\ref{sec:exp}.

\vspace{0.1cm}

\noindent{\textbf{Agent Harness Optimization}}. AI agents are undergoing a paradigm shift. Recent findings challenge the assumption that better models alone produce more reliable agents \cite{li2026agentharness,ning2026code}. Rather, 
improving the infrastructure layer around an agent, {\em i.e.}, the {\em agent harness}, can significantly enhance its performance \cite{li2026agentharness}, leading to growing attention to harness engineering \cite{lin2026agentic}, and emerging techniques for optimizing texts \cite{yuksekgonul2024textgrad} and harness \cite{metaharness2026} within agentic systems. Unlike 
time series agents that mostly rely on supervised fine-tuning or reinforcement learning \cite{guan2025timeomni,guan2026timeomni,wu2026timeart}, the proposed \method is, to our best knowledge, the first time series agent exposed to automatic harness optimization.



%% file: Sec/method.tex
\section{The Proposed Method}\label{sec:method}

\begin{figure*}[!t]
  \centering
  \includegraphics[width=\textwidth]{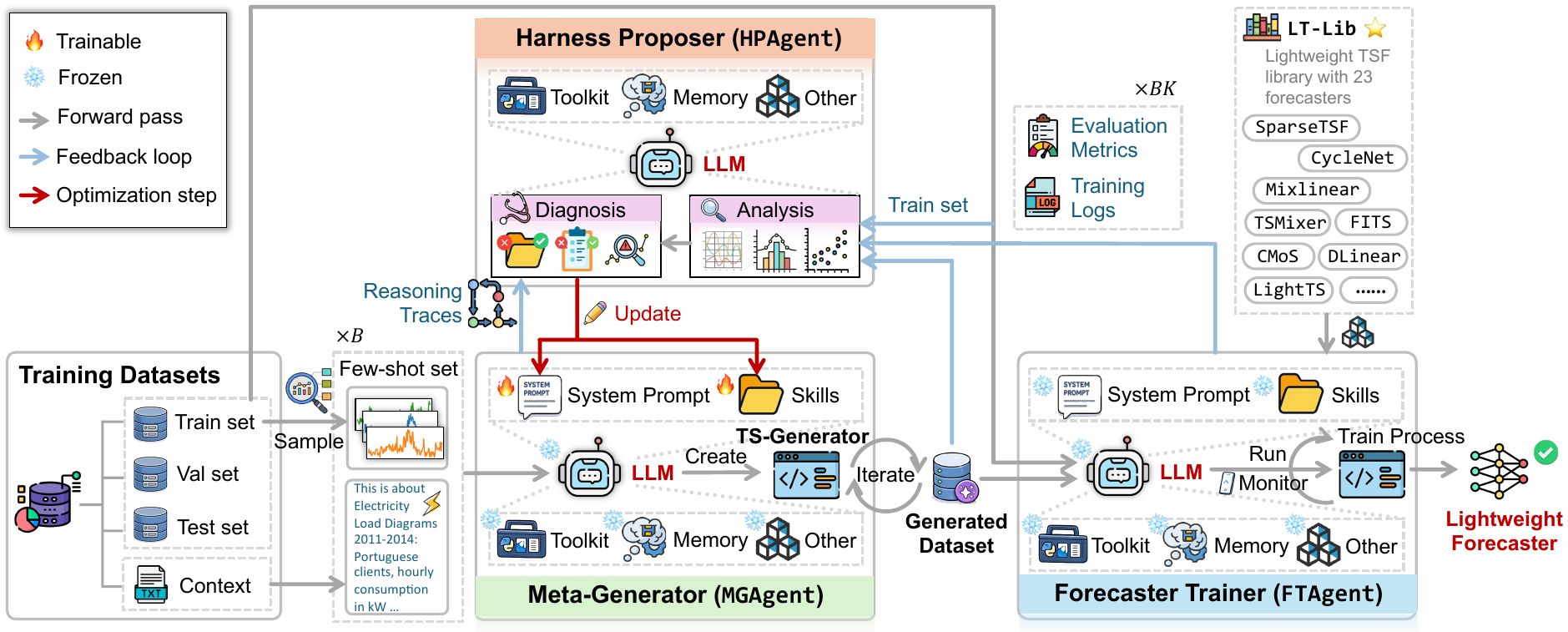}
  \caption{An illustration of the harness optimization framework of the proposed \method system.}\label{fig:method}
  \vspace{-0.3cm}
\end{figure*}

\subsection{Problem Statement}

Given a multivariate time series (MTS) $\mat{X}=[\mat{x}^{1}, ..., \mat{x}^{D}]^{\top}\in\mathbb{R}^{D\times T}$ within a {\em look-back window} of length $T$, where $\mat{x}^{d}\in\mathbb{R}^{T}$ ($1\le d \le D)$ is a univariate time series (UTS) of the $d$-th variate, the goal of TSF is to estimate the most likely values of the MTS at future $H$ time steps, {\em i.e.}, $\mat{\bar{Y}}\in\mathbb{R}^{D\times H}$, such that the difference between the estimation and the ground truth $\mat{Y}=\mat{X}_{T+1:T+H}\in\mathbb{R}^{D\times H}$ is minimized in terms of a metric, such as mean squared error (MSE).

In this work, we 
integrate multiple lightweight forecasters $\mathcal{F}=\{f_{1}, ..., f_{L}\}$, where $f_{l}:\mathbb{R}^{D\times T}\rightarrow\mathbb{R}^{D\times H}$ ($1\le l\le L$). We refer to $\mathcal{F}$ as \library (\underline{L}ightweight \underline{T}SF \underline{Lib}rary, \S\ref{sec.secondagent}). It is noteworthy that the forecasters in $\mathcal{F}$ are non-pre-trained due to their small capacities.

\vspace{0.1cm}

\noindent{\textbf{The Task}}. Given $\mathcal{F}$, a $K$-shot {\em support set} $\mathcal{D}_{\text{sup}}=\{(\mat{X}_{i}, \mat{Y}_{i})\}_{i=1}^{K}$, and a textual context description $\mathsf{C}$ about the target domain ({\em e.g.}, climatology, healthcare), the task is to 
train the forecasters in $\mathcal{F}$ with optimal forecasting errors on a test set. 
The task is challenging when $K$ is small, {\em i.e.}, in the few-shot setting, especially for 
the small forecasters in $\mathcal{F}$.

\subsection{Framework Overview}

To address the task, we propose \method, 
a multi-agent system that generates a sufficient dataset $\bar{\mathcal{D}}=\{(\mat{X}_{i}, \mat{Y}_{i})\}_{i=1}^{N'}$ with $N'\gg K$ based on the limited data in $\{\mathcal{D}_{\text{sup}}, \mathsf{C}\}$, and then splits $\bar{\mathcal{D}}$ into a training set $\bar{\mathcal{D}}_{\text{tr}}$ and a validation set $\bar{\mathcal{D}}_{\text{val}}$ for training the forecasters in $\mathcal{F}$.


Unlike existing generative models that focus on data realism \cite{timedp2025,verbalts2025,t2s2025}, \method generates $\bar{\mathcal{D}}$ to ensure that forecasters trained on $\bar{\mathcal{D}}_{\text{tr}}$ perform comparably to those trained on real data of the same size (as $|\bar{\mathcal{D}}_{\text{tr}}|$) in the target domain. This objective difference is crucial: standard generators cannot guarantee 
optimized data for training, and the resulting mismatch may lead to biased forecasters. In contrast, \method directly optimizes data generation for downstream forecasting performance, learning to produce training data that is better suited for building effective time series forecasters in the target domain.

To enable agent optimization, we adopt a meta-harness strategy \cite{metaharness2026} given recent findings on harness's significant impact on agent performance \cite{li2026agentharness}. 
Fig. \ref{fig:method} illustrates the optimization framework of \method, consisting of three key components: (1) a Meta-Generator (\firstagent, \S\ref{sec.firstagent}); (2) a Forecaster Trainer (\secondagent, \S\ref{sec.secondagent}); and (3) a Harness Proposer (\thirdagent, \S\ref{sec.thirdagent}). Among them, \thirdagent optimizes the Harness of \firstagent during the agent optimization process, which will be dropped at deployment and inference time (\S\ref{sec.inference}).

\subsection{The Harness Optimization Problem}\label{sec.optimization}


To enable \method to generate domain-specific training data, we construct a corpus of time series datasets $\mathcal{C}_{\text{har}}=\{\mathcal{D}^{m}, \mathsf{C}_{m}\}_{m=1}^{M}$ for Harness optimization, where $\mathcal{D}^{m}=\{(\mat{X}_{i}, \mat{Y}_{i})\}_{i=1}^{N_{m}}$ denotes the time series from the $m$-th domain and $\mathsf{C}_{m}$ its associated context. The corpus spans $M$ diverse domains to improve \method's cross-domain generalizability.


As illustrated in Fig. \ref{fig:method}, each dataset $\mathcal{D}^{m}$ is split into train/validation/test sets $\mathcal{D}_{\text{tr}}^{m}$, $\mathcal{D}_{\text{val}}^{m}$ and $\mathcal{D}_{\text{te}}^{m}$, representing the authentic data from the $m$-th domain. To simulate practical few-shot settings, $K$ samples from $\mathcal{D}_{\text{tr}}^{m}$ are used to construct $\mathcal{D}_{\text{sup}}^{m}$. \method then leverages $\{\mathcal{D}_{\text{sup}}^{m}, \mathsf{C}_{m}\}$ to generate $\{\bar{\mathcal{D}}_{\text{tr}}^{m}, \bar{\mathcal{D}}_{\text{val}}^{m}\}$. The objective is to minimize the performance gap between forecasters trained on $\{\mathcal{D}_{\text{tr}}^{m}, \mathcal{D}_{\text{val}}^{m}\}$ and those trained on $\{\bar{\mathcal{D}}_{\text{tr}}^{m}, \bar{\mathcal{D}}_{\text{val}}^{m}\}$, evaluated on the same test set $\mathcal{D}_{\text{te}}^{m}$.

Formally, let $\boldsymbol{\theta}$ be the trainable Harness in \method, $f_{l}^{m}$ ($\bar{f}_{l}^{m}$) be the forecaster trained using $\{\mathcal{D}_{\text{tr}}^{m}, \mathcal{D}_{\text{val}}^{m}\}$ ($\{\bar{\mathcal{D}}_{\text{tr}}^{m}, \bar{\mathcal{D}}_{\text{val}}^{m}\}$), the Harness optimization problem is
\begin{equation}\label{eq.loss}
\begin{aligned}
\min_{\boldsymbol{\theta}}{\mathbb{E}_{1\le m\le M, 1\le l\le L}\big(\delta(\omega(f_{l}^{m}), \omega(\bar{f}_{l}^{m}))\big)}
\end{aligned}
\end{equation}
where $\omega(\cdot)$ is a metric that evaluates forecasting errors on $\mathcal{D}_{\text{te}}^{m}$, and $\delta(\cdot, \cdot)$ is a measure of difference between two metrics.

\subsection{Meta-Generator (\firstagent)}\label{sec.firstagent}

Given the few-shot set $\{\mathcal{D}_{\text{sup}}^{m}, \mathsf{C}_{m}\}$, \firstagent seeks to generate $\bar{\mathcal{D}}^{m}$, that is
\begin{equation}\label{eq.firstagnet}
\begin{aligned}
\bar{\mathcal{D}}^{m} = \firstagent(\mathcal{D}_{\text{sup}}^{m}, \mathsf{C}_{m})
\end{aligned}
\end{equation}
which will be split into $\{\bar{\mathcal{D}}_{\text{tr}}^{m}, \bar{\mathcal{D}}_{\text{val}}^{m}\}$ by \secondagent (\S\ref{sec.secondagent}). 
Here, $\mathsf{C}_{m}$ provides semantic cues that guide \firstagent to generate domain-compliant signals using methods pertinent to the $m$-th domain.


As shown in Fig. 2, \firstagent is centered on a (replaceable) LLM. Rather than generating MTS directly, the LLM uses its Harness to create a \texttt{TS-Generator} program that integrates domain-specific knowledge, rules, and models pertinent to the $m$-th domain, circumventing LLM's limited capability in direct time series inference \cite{merrill2024language}. Instead, \firstagent exploits the LLM’s strengths in planning, reasoning, and coding to orchestrate more suitable tools for time series inference, hence the ``Meta-'' prefix.

Upon receiving $\{\mathcal{D}_{\text{sup}}^{m}, \mathsf{C}_{m}\}$, \firstagent (1) analyzes the MTS in $\mathcal{D}_{\text{sup}}^{m}$, 
(2) creates \texttt{TS-Generator} 
to generate $\bar{\mathcal{D}}^{m}$, and (3) performs quality checks. It accepts $\bar{\mathcal{D}}^{m}$ if all checks pass. Otherwise, it goes to step (2) to revise \texttt{TS-Generator}.


In this process, the Harness exposes several components to the LLM, including a {\em system prompt}, {\em skills}, a {\em toolkit}, {\em long-term memory}, and other execution {\em middleware}. The system prompt defines \firstagent's behavior, goals, constraints, and tool-usage rules. Skills \cite{li2026skillsbench} are reusable modules containing instructions, metadata, and tool bindings for the time series generation task in Eq.~\eqref{eq.firstagnet}. The toolkit provides APIs for tools such as Python and web search, while the long-term memory stores information about generated data to support iterative review and refinement in steps (2) and (3).

In the Harness, the system prompt and skills define key behavior of \firstagent. Thus, 
we freeze the LLM and use them as the trainable parameters $\boldsymbol{\theta}$. Instead of Harness engineering \cite{li2026agentharness}, $\boldsymbol{\theta}$ is edited by \thirdagent (\S\ref{sec.thirdagent}) in an outer loop to automatically optimize Eq.~\eqref{eq.loss}.

\subsection{Forecaster Trainer (\secondagent)}\label{sec.secondagent}

\secondagent splits $\bar{\mathcal{D}}^{m}$ into $\{\bar{\mathcal{D}}_{\text{tr}}^{m}, \bar{\mathcal{D}}_{\text{val}}^{m}\}$ according to the partition sizes of $\{\mathcal{D}_{\text{tr}}^{m}, \mathcal{D}_{\text{val}}^{m}\}$, orchestrates computational resources to train the forecasters in $\mathcal{F}$, and evaluates them on $\mathcal{D}_{\text{te}}^{m}$, producing trained forecasters, evaluation metrics, and training reports:
\begin{equation}\label{eq.secondagent}
\begin{aligned}
&\{\{f_{l}^{m}, \bar{f}_{l}^{m}, \omega(f_{l}^{m}), \omega(\bar{f}_{l}^{m})\}_{l=1}^{L}, \mathsf{R}\}\\
&= \secondagent(\{\mathcal{D}_{\text{tr}}^{m}, \mathcal{D}_{\text{val}}^{m}, \mathcal{D}_{\text{te}}^{m}\}, \{\bar{\mathcal{D}}_{\text{tr}}^{m}, \bar{\mathcal{D}}_{\text{val}}^{m}\})
\end{aligned}
\end{equation}
where $\omega(\cdot)$ is the evaluation metric such as MSE and MAE, and $\mathsf{R}$ denotes the training reports.

\vspace{0.1cm}

\noindent{\textbf{Lightweight TSF Library (\library)}}. To construct $\mathcal{F}$, we collect 23 SOTA lightweight forecasters proposed from 2022 to 2026, including linear-layer models 
({\em e.g.}, \texttt{MixLinear} \cite{li2024mixlinear}), MLP-based forecasters 
({\em e.g.}, \texttt{TSMixer} \cite{chen2023tsmixer}), and frequency-domain models 
({\em e.g.}, \texttt{FITS} \cite{xu2024fits}). Their sizes, as listed in Appendix~\ref{app:forecasters}, 
are much smaller than SOTA Transformer-based forecasters ({\em e.g.}, \texttt{PatchTST} \cite{nie2023patchtst}: $\sim$3.2M) and TSFMs ({\em e.g.}, \texttt{Chronos} \cite{ansari2024chronos}: $\sim$700M). 
We compile them in \library with a unified interface to facilitate training calls in \secondagent.


As shown in Fig. \ref{fig:method}, \secondagent composes programs to train each forecaster with a grid search of hyperparameters. It organizes training jobs -- {\em i.e.}, (forecaster, hyperparameter, dataset) triplets -- into a queue and assigns available GPUs for maximally parallel execution. During training, it monitors processes, resolves errors when encountered, and recovers interrupted jobs without human intervention. This procedure ends when the queue is emptied.


\secondagent's Harness 
uses the same kind of components as \firstagent's, with additional access to \library. Unlike \firstagent, however, its Harness is not optimized, since the training process is relatively standard and has less impact on the resulting forecasters than the generated dataset $\bar{\mathcal{D}}^{m}$.



\subsection{Harness Proposer (\thirdagent)}\label{sec.thirdagent}

\thirdagent is the meta-harness \cite{metaharness2026} that configures the Harness of \firstagent to optimize Eq.~\eqref{eq.loss}. Upon receiving $\{\omega(f_{l}^{m}), \omega(\bar{f}_{l}^{m})\}_{l=1}^{L}$ from \secondagent, 
\thirdagent evaluates Eq.~\eqref{eq.loss} using a hinge-loss based measure:
\begin{equation}\label{eq.distance}
\begin{aligned}
\delta(\omega(f_{l}^{m}), \omega(\bar{f}_{l}^{m}))=\max{\Big\{\frac{\omega(\bar{f}_{l}^{m}) - \omega(f_{l}^{m})}{\omega(f_{l}^{m})}, 0\Big\}}
\end{aligned}
\end{equation}
which induces a penalty when the forecasting error $\omega(\bar{f}_{l}^{m})$ of the forecaster trained on the generated $\bar{\mathcal{D}}_{\text{tr}}$ is larger than $\omega(f_{l}^{m})$ of the forecaster trained on the authentic dataset $\mathcal{D}_{\text{tr}}$, suggesting $\bar{\mathcal{D}}_{\text{tr}}$ needs improvements. 
Otherwise, Eq.~\eqref{eq.distance} won't induce any penalty when $\bar{f}_{l}^{m}$ performs better than $f_{l}^{m}$.


As in Fig. \ref{fig:method}, \thirdagent oversees the entire pipeline through three stages: (1) {\em Self-Planned Analysis}, which collects evidence on \firstagent and \secondagent performance, including the loss in Eq.~\eqref{eq.loss}, statistical differences between $\mathcal{D}_{\text{tr}}$ and $\bar{\mathcal{D}}_{\text{tr}}$ ({\em e.g.}, MMD), and training issues from logs $\mathsf{R}$; (2) {\em Diagnosis}, which identifies potential causes of performance degradation from \firstagent's reasoning traces, system prompt, and skills, then determines how to update the Harness $\boldsymbol{\theta}$ of \firstagent; and (3) {\em Update}, which edits $\boldsymbol{\theta}$. Effectively, \thirdagent serves as the optimizer of Eq.~\eqref{eq.loss}, as defined by its system prompt.


Accordingly, \thirdagent's lifecycle is the entire optimization process with multiple epochs of updating $\boldsymbol{\theta}$, whereas \firstagent and \secondagent each functions within a single epoch. \thirdagent therefore relies more heavily on long-term memory, which stores dataset snapshots, evaluation metrics, training logs, analysis/diagnosis results, $\boldsymbol{\theta}$ update logs, {\em etc.}, across epochs. This memory enables rollback of harmful updates and allows \thirdagent to output the best $\boldsymbol{\theta}$ 
at the end of optimization.

\subsection{Training and Inference}\label{sec.inference}

\noindent{\textbf{Harness Training}}. We summarize the optimization process in Appendix~\ref{app:algorithm}. In every epoch, we draw $K\sim\text{Uniform}([10, 50])$. Notably, 
instead of feeding the $K$-shot support set $\{\mathcal{D}_{\text{sup}}^{m}, \mathsf{C}_{m}\}$ to \method individually, we sample a batch of $B$ datasets from the corpus $\mathcal{C}_{\text{har}}$ to extract $B$ support sets, and let \method  process them simultaneously. This extension allows \thirdagent to evaluate generalizability across datasets and mitigate overfitting to any specific dataset.

\vspace{0.1cm}


\noindent{\textbf{Inference}}. After optimization, \thirdagent outputs the best $\boldsymbol{\theta}^{*}$. 
Then \thirdagent is discarded. The Harnesses of \firstagent (including $\boldsymbol{\theta}^{*}$) and \secondagent (including \library) are retained, while their LLMs 
are removed. For a new target domain $g$, these Harnesses are attached to user-selected LLM APIs to instantiate \firstagent and \secondagent. Given $K$-shot $\mathcal{D}_{\text{sup}}^{g}$ and context $\mathsf{C}_{g}$, the agents execute the forward process in Fig. \ref{fig:method} to produce the best trained forecaster $\bar{f}_{*}^{g}$. At TSF stage, only $\bar{f}_{*}^{g}$ needs to be maintained, while \firstagent and \secondagent are dropped, as shown in Fig. \ref{fig.intro}(c).

Therefore, the entire downstream deployment and inference process can be accomplished using low-cost devices capable of running the Harnesses and the \library framework, making it more efficient than relying on GPU-intensive TSFMs.

%% file: Sec/exp.tex
\input{Tab/main_results}

\section{Experiments}
\label{sec:exp}

\begin{figure*}[t]
  \centering
  \includegraphics[width=\textwidth]{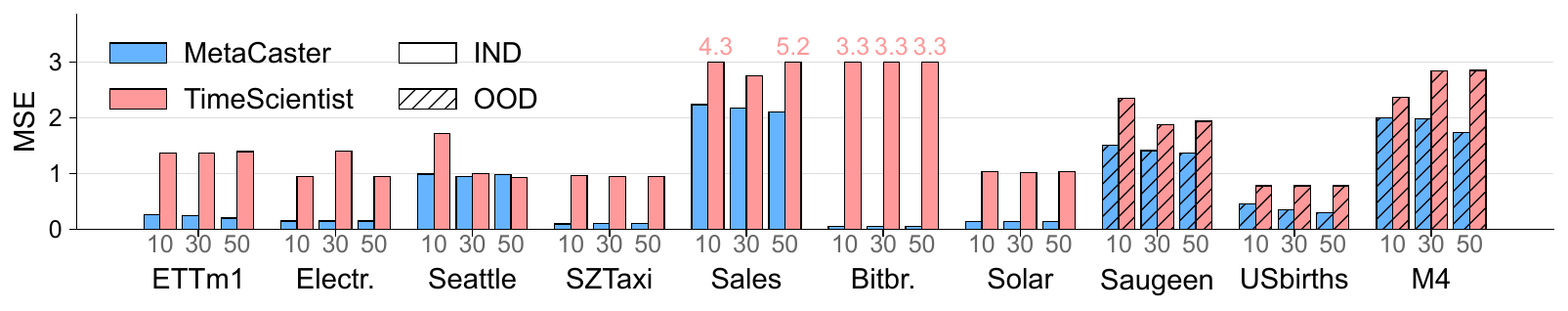}
  \caption{Comparing the selected (trained) forecasters of agent pipelines \method and \texttt{TimeScientist}.}
  \label{fig:tsci-bars}
  \vspace{-0.2cm}
\end{figure*}

\subsection{Experimental Setup}\label{ssec:exp-setup}


\noindent{\textbf{Datasets}. To evaluate \method, we employ the GIFT-Eval benchmark \cite{gifteval2024}, previsouly adopted for training TSFMs \cite{liu2024moirai}. We use 18 datasets across 9 domains ({\em e.g.}, energy, weather, traffic), split into a train corpus of 8 datasets ({\em i.e.}, $\mathcal{C}_{\text{har}}$ in \S\ref{sec.optimization}), an {\em in-domain (IND)} test corpus (7 datasets), and an {\em out-of-domain (OOD)} test corpus (3 datasets). The IND corpus shares domains with $\mathcal{C}_{\text{har}}$, while the OOD corpus covers unseen domains to evaluate generalization. All test datasets are disjoint from $\mathcal{C}_{\text{har}}$. Each dataset $\mathcal{D}^{m}$ has a textual context $\mathsf{C}_{m}$. Time series are split chronologically into 80\%/10\%/10\% train/validation/test sets. Following standard protocol \cite{nie2023patchtst}, the look-back window $T$ is set as 336, the prediction horizon $H$ is 192. Full dataset details and approaches to avoid data leakage are deferred to Appendix \ref{app:bench}.


\vspace{0.1cm}

\noindent{\textbf{Baselines}}. We compare \method with the most relevant SOTA methods, including {\em time series generation models}: (1) \texttt{TimeVAE} \cite{desai2021timevae}, (2) \texttt{DiffTS} \cite{yuan2024diffusionts}, (3) \texttt{T2S} \cite{t2s2025}, (4) \texttt{TimeDP} \cite{timedp2025}, (5) \texttt{VerbalTS} \cite{verbalts2025}; {\em time series augmentation techniques} \cite{iwana2021empirical,luo2023time}: (6) \texttt{Repeat}, (7) \texttt{Bootstrap}, (8) \texttt{Jitter}, (9) \texttt{MagWarp}; {\em pre-trained TSFMs}: (10) \texttt{Chronos} \cite{ansari2024chronos}, (11) \texttt{Moirai} \cite{liu2024moirai}, (12) \texttt{VisionTS} \cite{visionts2024}, (13) \texttt{Time-LLM} \cite{jin2024timellm}; and a SOTA {\em agent pipeline}: (14) \texttt{TimeScientist} \cite{zhao2025timeseriesscientist}. 
Details about these methods are in Appendix~\ref{app:baselines}.

\vspace{0.1cm}

\noindent{\textbf{Settings}}. We draw $K\in\{10, 30, 50\}$ samples from each test dataset 
to constitute $\mathcal{D}_{\text{sup}}^{m}$, which is fed with $\mathsf{C}_{m}$ to the compared methods. Generation models and augmentation methods use $\{\mathcal{D}_{\text{sup}}^{m}, \mathsf{C}_{m}\}$ to produce $\bar{\mathcal{D}}^{m}$ of size $N_{m}'$, which is used to train the forecasters in our \library. 
TSFMs use $\mathcal{D}_{\text{sup}}^{m}$ for fine-tuning. \texttt{TimeScientist} uses $\mathcal{D}_{\text{sup}}^{m}$ to directly train forecasters as it cannot generate time series.

For our \method, we set batch size $B=8$, and use \texttt{GPT-5.4} \cite{metaharness2026} as the default LLM. We also test different LLMs in our ablation study (Table~\ref{tab:ablation}). For all methods that can generate $\bar{\mathcal{D}}^{m}$, the size $N_{m}'$ is set to $|\mathcal{D}_{\text{tr}}^{m}|+|\mathcal{D}_{\text{val}}^{m}|$. 
Following \cite{nie2023patchtst,tan2024language}, we use MSE and MAE to evaluate the TSF performance.

\subsection{Experimental Results}
\label{ssec:exp-main}



Table \ref{tab:main} reports MSE results comparing generation models, augmentation methods, and \method. We randomly hold out 3 forecasters from \library and evaluate the average performance over the remaining 20; the held-out forecasters are used to assess \method's generalization to unseen forecasters (Fig. \ref{fig.boxplot}). MAE results are provided in Appendix \ref{app:mae}, and Appendix \ref{app:topk} reports the best-performing forecaster among the 20 as trained by the compared methods.

\begin{figure}[!t]
\centering
\includegraphics[width=\columnwidth]{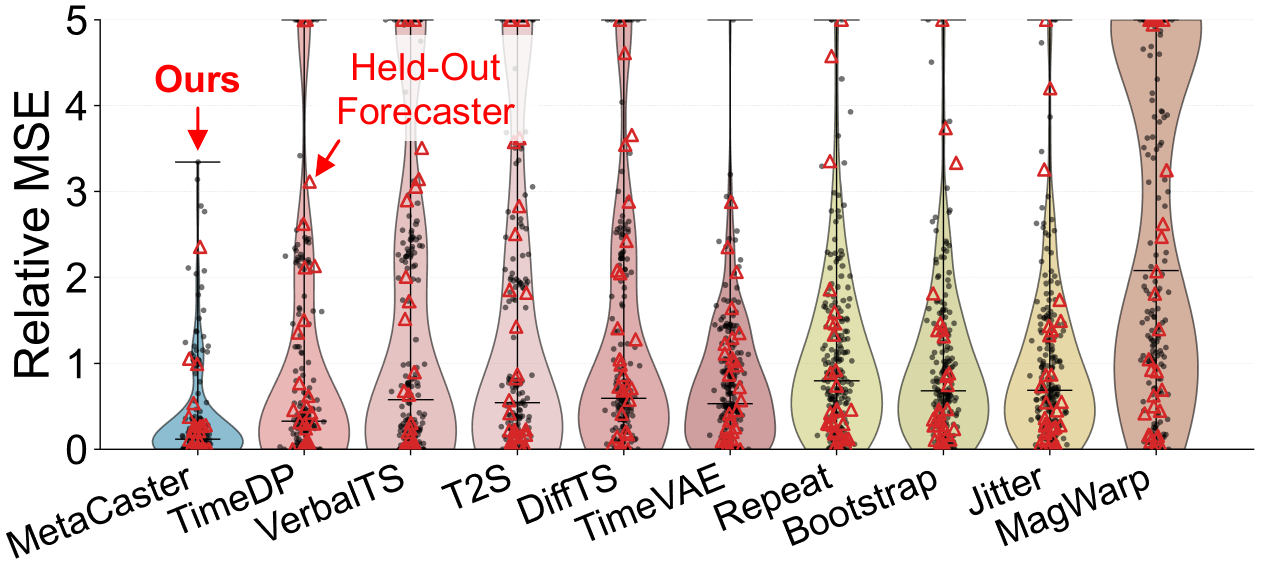}
\caption{Performance distribution. Each dot represents the performance of a forecaster $\bar{f}_{l}^{m}$ on a test set $\mathcal{D}_{\text{te}}^{m}$, covering all forecasters in \library and all datasets in both of IND and OOD corpora.}\label{fig.boxplot}
\vspace{-0.2cm}
\end{figure}


Table \ref{tab:main} also reports results of training forecasters with $K$-shot $\mathcal{D}_{\text{sup}}^{m}$ and full set $\mathcal{D}_{\text{tr}}^{m}$, serving as lower and upper performance references, respectively. From Table \ref{tab:main}, we highlight several observations: (1) \method outperforms the generation/augmentation baselines in most cases, demonstrating the benefit of optimizing data generation for forecasting; (2) \method's performance improves with larger $K$, showing effective few-shot utilization; (3) When $K\ge 30$, \method approaches or even surpasses $\mathcal{D}_{\text{tr}}^{m}$, suggesting that raw data may be noisy and optimized data can improve training; (4) In the extreme case when $K=10$, \method remains competitive; and (5) Despite increased difficulty, \method generalizes well to OOD datasets, where it generally outperforms baselines.


Fig. \ref{fig.boxplot} shows the MSE distribution of individual forecasters trained by the models in Table \ref{tab:main}, normalized by the upper reference ($\mathcal{D}_{\text{tr}}^{m}$) ({\em i.e.}, Eq.~\eqref{eq.distance}). \method produces more high-quality forecasters with lower variance. Moreover, it generalizes better to held-out forecasters than the baselines. 
Fig. \ref{fig:tsci-bars} compares \method with \texttt{TimeScientist} under $K \in \{10, 30, 50\}$. Without data generation capability, \texttt{TimeScientist} struggles to train generalizable forecasters and its performance does not scale with $K$, indicating that a pure training pipeline is insufficient. Appendix \ref{app:tsci-lf} reports results when \texttt{TimeScientist} uses \library, yielding consistent conclusions.

\vspace{0.1cm}


\noindent{\textbf{Comparing with TSFMs}}.\label{ssec:exp-foundation}
Fig. \ref{fig.tsfm} compares MetaCaster with TSFMs on the Solar dataset under $K=30$ in terms of performance {\em vs.} cost. After deployment, \method trains and selects a lightweight forecaster for inference and incurs no further agent overhead -- here, \texttt{MixLinear} (243 parameters) is selected at runtime. In contrast, TSFMs remain expensive from fine-tuning through inference. At comparable performance, \method achieves up to $10^{3}\times$ lower latency and $10^{4}\times$ fewer parameters than TSFMs.

\begin{figure}[!t]
\centering
\includegraphics[width=0.84\columnwidth]{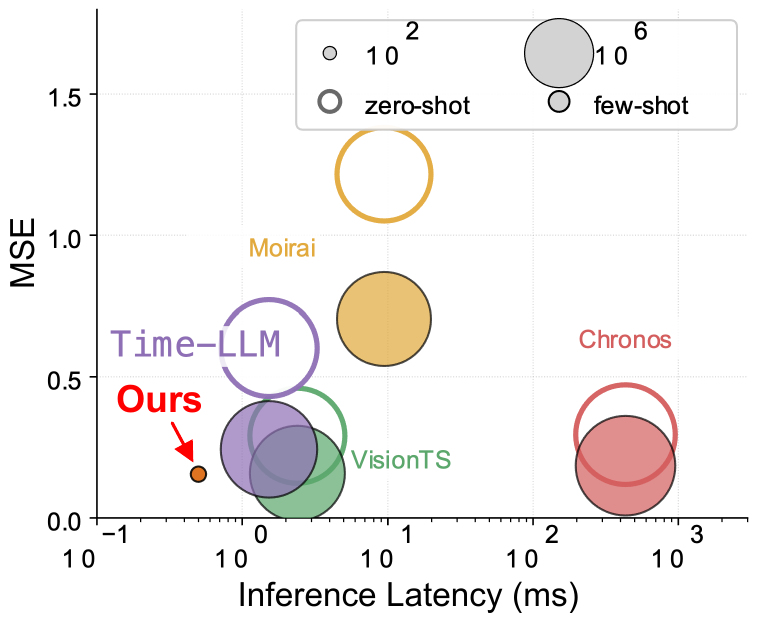}
\caption{Comparing \method with TSFMs on Solar dataset. Full results are in Appendix~\ref{app:foundation-full}.}\label{fig.tsfm}
\vspace{-0.25cm}
\end{figure}

\input{Tab/ablation_design}

\begin{figure*}[t]
  \centering
  \includegraphics[width=\textwidth]{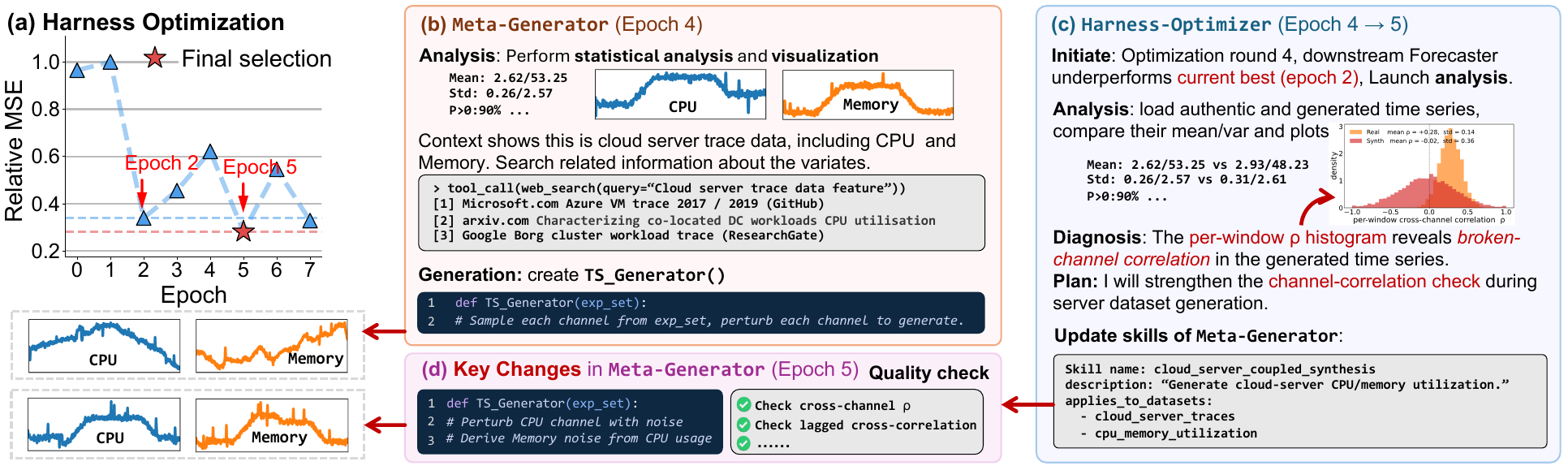}
  \caption{An illustration of Harness optimization by \method: (a) hinge loss (Eq.~\eqref{eq.distance}) during optimization iterations. (b)-(d) key traces of \firstagent and \thirdagent from Epoch 4 to 5 on alibaba\_cluster\_2018 dataset.}\label{fig:casestudy}
\end{figure*}

\vspace{0.1cm}


\noindent{\textbf{Ablation Study}}. Table \ref{tab:ablation} presents an ablation study using averaged MSE with $K=30$, with \method as our original model. In (a), we study the effect of forecasting-oriented optimization in Eq.~\eqref{eq.loss} by replacing Eq.~\eqref{eq.loss} with MMD and Wasserstein distances to directly align the generated set $\bar{\mathcal{D}}_{\text{tr}}^{m}$ with the authentic set $\mathcal{D}_{\text{tr}}^{m}$. In (b), we remove contextual cues $\mathsf{C}_{m}$ to evaluate their contribution. In (c), we examine the impact of different LLMs.


From Table \ref{tab:ablation}(a), minimizing data distribution discrepancy generally degrades performance, as it is not directly aligned with the forecasting objective. Table \ref{tab:ablation}(b) shows that contextual cues $\mathsf{C}_{m}$ are crucial, as they guide \firstagent in selecting domain-relevant knowledge for generating time series. Interestingly, Table \ref{tab:ablation}(c) shows that different LLMs yield comparable results in many cases, suggesting the Harness -- rather than the LLM backbone -- is the key factor in \method, consistent with prior findings \cite{li2026agentharness}. Although \texttt{GPT-5.3-Codex} wins in some cases, it is unstable on datasets such as ETTm1 and USbirths, leading to worse overall results; thus, we adopt \texttt{GPT-5.4} as the default LLM for its consistent performance.

\vspace{0.1cm}

\noindent{\textbf{Further Analysis}}. We have evaluated computational efficiency, token usage, robustness of performance (using standard deviation), performance change {\em w.r.t.} $K$, performance of top-ranked forecasters, visualization of generated time series and data distribution, which are in Appendix \ref{app:budget}, \ref{app:further}.

\subsection{Case Study}\label{ssec:exp-case}


Fig. \ref{fig:casestudy}(a) shows the evolution of hinge loss (Eq.~\eqref{eq.distance}) over 8 Harness optimization epochs, where \method converges quickly and selects the final Harness from epoch 5. Fig \ref{fig:casestudy}(b)-(d) illustrates \firstagent and \thirdagent traces from epoch 4 to 5 on the alibaba\_cluster\_2018 dataset in $\mathcal{C}_{\text{har}}$. \firstagent (Fig \ref{fig:casestudy}(b)) analyzes the few-shot examples $\mathcal{D}_{\text{sup}}^{m}$, inspects statistics, retrieves domain knowledge about the two variates CPU and memory, and constructs a \texttt{TS\_Generator}. \thirdagent (Fig. \ref{fig:casestudy}(c)) detects performance degradation at epoch 4 via memory, then performs Analysis and Diagnosis. The Analysis uses analytical tools pertinent to the issue identified in the reasoning traces of \firstagent and the training logs of \secondagent. The Diagnosis identifies broken inter-variate correlation as the root cause. It then updates \firstagent's skills, leading to an improved \texttt{TS\_Generator} that strengthens the CPU-memory correlation and introduces an addition of consistency checks. Consequently, epoch 5 produces correlated time series that are more compliant with the few-shot examples than epoch 4. This example illustrates a simple Harness update. In fact, \method's optimization is more complex and it deals with $B=8$ datasets in $\mathcal{C}_{\text{har}}$ concurrently. The final output is a trained lightweight forecaster rather than the generated time series.

%% file: Tab/main_results.tex
\begin{table*}[!t]
  \centering
  \footnotesize
  \resizebox{\textwidth}{!}{%
  \begin{tabular}{c l | c | c c c c c | c c c c | c c}
    \toprule
    & \multirow{2}{*}{\textbf{Dataset}}
      & \textbf{Ours}
      & \multicolumn{5}{c|}{\textbf{Generation Models}}
      & \multicolumn{4}{c|}{\textbf{Augmentation Methods}}
      & \multicolumn{2}{c}{\textbf{References}} \\
    & & \method & \texttt{TimeDP} & \texttt{VerbalTS} & \texttt{T2S} & \texttt{DiffTS} & \texttt{TimeVAE} & \texttt{Repeat} & \texttt{Bootstrap} & \texttt{Jitter} & \texttt{MagWarp} & $\mathcal{D}_{\text{sup}}^{m}$ & $\mathcal{D}_{\text{tr}}^{m}$ \\
    \midrule
    \multirow{7}{*}{\rotatebox{90}{\textbf{IND ($K=10$)}}}
      & ETTm1       & \best{0.376} & 1.049 & 1.004 & 0.922 & 1.154 & 0.749 & 0.781 & 0.748 & \second{0.746} & 1.919 & 0.687 & 0.316 \\
      & Electricity & \second{0.300} & \best{0.288} & 1.061 & 1.150 & 0.365 & 0.328 & 0.380 & 0.379 & 0.375 & 7.057 & 0.721 & 0.121 \\
      & Seattle     & \second{1.522} & \best{1.420} & 1.800 & 1.792 & 1.643 & 1.759 & 1.955 & 1.849 & 1.925 & 43.24 & 2.341 & 1.024 \\
      & SZTaxi      & \best{0.104} & 0.124 & 0.116 & 0.118 & \second{0.111} & 0.114 & 0.134 & 0.118 & 0.123 & 1.122 & 0.187 & 0.096 \\
      & Sales       & 2.616 & \second{2.587} & \best{2.419} & 2.758 & 4.323 & 2.795 & 4.175 & 2.949 & 2.946 & 4.321 & 3.004 & 2.927 \\
      & Bitbrains   & \best{0.068} & \second{0.079} & 0.098 & 0.094 & 0.121 & 0.109 & 0.197 & 0.149 & 0.159 & 0.826 & 0.152 & 0.150 \\
      & Solar       & \best{0.158} & \second{0.208} & 0.513 & 0.552 & 0.238 & 0.237 & 0.278 & 0.272 & 0.272 & 0.446 & 0.440 & 0.157 \\
    \cmidrule(lr){2-14}
    \multirow{3}{*}{\rotatebox{90}{\textbf{OOD}}}
      & Saugeen     & 1.960 & \second{1.492} & \best{1.474} & 1.578 & 2.281 & 2.705 & 3.119 & 2.887 & 2.908 & 3.893 & 1.859 & 1.183 \\
      & USbirths    & \second{0.702} & 2.690 & 1.501 & 1.540 & 1.775 & \best{0.607} & 0.752 & 0.715 & 0.714 & 12.08 & 1.204 & 0.250 \\
      & M4\textsuperscript{$\dagger$} & \best{2.093} & \second{2.178} & 2.331 & 2.444 & 2.688 & 2.993 & 3.250 & 3.101 & 3.057 & 8.654 & 2.628 & 2.330 \\
    \midrule
    \multirow{7}{*}{\rotatebox{90}{\textbf{IND ($K=30$)}}}
      & ETTm1       & \best{0.345} & 1.049 & 1.027 & 0.910 & 1.211 & 0.684 & 0.714 & 0.664 & \second{0.662} & 0.943 & 0.602 & 0.316 \\
      & Electricity & \best{0.226} & 0.287 & 1.055 & 1.140 & 0.363 & \second{0.268} & 0.305 & 0.287 & 0.290 & 1.644 & 0.657 & 0.121 \\
      & Seattle     & \best{1.177} & \second{1.414} & 1.806 & 1.777 & 1.673 & 2.158 & 1.738 & 1.656 & 1.622 & 21.15 & 1.785 & 1.024 \\
      & SZTaxi      & \second{0.114} & 0.121 & 0.116 & 0.115 & \best{0.112} & 0.156 & 0.162 & 0.141 & 0.135 & 1.626 & 0.150 & 0.096 \\
      & Sales       & \best{2.362} & 2.479 & \second{2.455} & 2.942 & 4.439 & 2.598 & 3.603 & 3.131 & 3.158 & 3.422 & 2.611 & 2.927 \\
      & Bitbrains   & 0.125 & \best{0.084} & 0.097 & \second{0.095} & 0.130 & 0.114 & 0.463 & 0.469 & 0.489 & 0.777 & 0.137 & 0.150 \\
      & Solar       & \best{0.152} & \second{0.220} & 0.515 & 0.576 & 0.240 & 0.234 & 0.254 & 0.252 & 0.248 & 0.356 & 0.379 & 0.157 \\
    \cmidrule(lr){2-14}
    \multirow{3}{*}{\rotatebox{90}{\textbf{OOD}}}
      & Saugeen     & \best{1.464} & 1.495 & \second{1.480} & 1.593 & 2.286 & 2.019 & 2.419 & 2.266 & 2.277 & 2.620 & 1.547 & 1.183 \\
      & USbirths    & 0.533 & 1.980 & 1.509 & 1.536 & 1.732 & \best{0.474} & 0.546 & \second{0.524} & 0.525 & 5.107 & 0.803 & 0.250 \\
      & M4\textsuperscript{$\dagger$} & \best{2.112} & \second{2.248} & 2.394 & 2.522 & 2.663 & 3.402 & 3.126 & 3.109 & 3.169 & 10.321 & 2.950 & 2.330 \\
    \midrule
    \multirow{7}{*}{\rotatebox{90}{\textbf{IND ($K=50$)}}}
      & ETTm1       & \best{0.267} & 1.063 & 1.035 & 0.931 & 1.284 & 0.548 & 0.582 & 0.556 & \second{0.546} & 0.592 & 0.456 & 0.316 \\
      & Electricity & \best{0.191} & 0.279 & 1.053 & 1.197 & 0.352 & \second{0.227} & 0.244 & 0.234 & 0.233 & 1.287 & 0.472 & 0.121 \\
      & Seattle     & \best{1.125} & \second{1.423} & 1.819 & 1.779 & 1.732 & 2.810 & 1.537 & 1.462 & 1.497 & 12.78 & 1.533 & 1.024 \\
      & SZTaxi      & \best{0.110} & 0.120 & 0.114 & \second{0.113} & \second{0.113} & 0.318 & 0.230 & 0.222 & 0.216 & 0.352 & 0.143 & 0.096 \\
      & Sales       & 2.835 & 2.605 & \best{2.474} & 2.852 & 4.459 & 2.683 & 2.778 & 2.590 & \second{2.560} & 3.039 & 2.501 & 2.927 \\
      & Bitbrains   & 0.119 & \second{0.097} & 0.099 & \best{0.095} & 0.124 & 0.155 & 0.929 & 0.976 & 0.987 & 1.214 & 0.119 & 0.150 \\
      & Solar       & \best{0.149} & 0.234 & 0.520 & 0.553 & 0.250 & 0.547 & 0.234 & 0.231 & \second{0.229} & 0.333 & 0.296 & 0.157 \\
    \cmidrule(lr){2-14}
    \multirow{3}{*}{\rotatebox{90}{\textbf{OOD}}}
      & Saugeen     & \best{1.415} & \second{1.464} & 1.482 & 1.591 & 2.291 & 1.767 & 2.462 & 2.380 & 2.371 & 2.528 & 1.515 & 1.183 \\
      & USbirths    & 0.424 & 2.366 & 1.509 & 1.531 & 1.687 & \best{0.367} & 0.396 & 0.380 & \second{0.376} & 5.787 & 0.574 & 0.250 \\
      & M4\textsuperscript{$\dagger$} & \best{1.916} & \second{2.157} & 2.455 & 2.444 & 2.896 & 3.279 & 3.330 & 3.300 & 3.282 & 12.210 & 2.560 & 2.330 \\
    \midrule
    & Wins (of 30) & 19 & 3 & 3 & 1 & 1 & 3 & 0 & 0 & 0 & 0 & -- & -- \\
    \bottomrule
  \end{tabular}%
  }
  \caption{TSF performance comparison on IND and OOD corpora for $K\in\{10, 30, 50\}$ in terms of MSE. Lower MSE is better. \best{Red} (\second{blue}) values indicate the best (second-best) MSE per row. \textsuperscript{$\dagger$}M4 uses instance-normalized MSE to address large distribution shifts. MAE results are available in Appendix~\ref{app:mae}.}
  \label{tab:main}
  \vspace{-0.3cm}
\end{table*}

%% file: Tab/ablation_design.tex
\begin{table*}[t]
  \centering
  \small
  \resizebox{\textwidth}{!}{%
  \begin{tabular}{l | c c c c c c c | c c c | c}
    \toprule
    \textbf{Dataset} ($\rightarrow$)
      & \multicolumn{7}{c|}{\textbf{IND Corpus}}
      & \multicolumn{3}{c|}{\textbf{OOD Corpus}}
      & \\
    \textbf{Method} ($\downarrow$)
      & ETTm1 & Electricity & Seattle & SZTaxi & Sales & Bitbrains & Solar
      & Saugeen & USbirths & M4
      & \textbf{Overall} \\
    \midrule
      \method
        & \best{0.345} & 0.226 & \second{1.177} & \second{0.114} & \second{2.362} & \best{0.125} & 0.152 & \best{1.464} & \second{0.533} & \best{2.112} & \best{0.267} \\
    \midrule
      (a) Loss$\rightarrow$MMD
        & 0.708 & \best{0.157} & 1.204 & 0.150 & 3.581 & 0.575 & 0.254 & 2.339 & \best{0.403} & 2.556 & 0.764 \\
      (a) Loss$\rightarrow$Wasserstein
        & 0.702 & 0.284 & 1.734 & 0.152 & 3.553 & 0.483 & 0.254 & 2.406 & 0.535 & 3.105 & 0.940 \\
    \midrule
      (b) Remove context $\mathsf{C}_{m}$
        & \second{0.386} & 0.293 & 1.425 & \second{0.144} & 2.595 & 0.158 & 0.281 & 2.004 & 0.534 & 2.312 & 0.521 \\
    \midrule
      (c) LLM$\rightarrow$\texttt{Gemini-3.1-Pro}
        & 0.430 & 0.226 & \second{1.177} & \second{0.114} & \second{2.362} & 0.120 & 0.151 & 1.396 & \second{0.533} & \second{2.129} & \second{0.288} \\
      (c) LLM$\rightarrow$\texttt{Claude-Opus-4.7}
        & 0.494 & 0.226 & 1.189 & \second{0.114} & \second{2.362} & 0.151 & \second{0.150} & 1.491 & 0.541 & \second{2.129} & 0.321 \\
      (c) LLM$\rightarrow$\texttt{Qwen3.5-122B-A10B}
        & 0.494 & 0.223 & \second{1.177} & \second{0.114} & \second{2.362} & 0.199 & 0.152 & 1.729 & \second{0.533} & \second{2.129} & 0.366 \\
      (c) LLM$\rightarrow$\texttt{GPT-5.3-Codex}
        & 0.557 & \second{0.209} & \best{1.115} & \best{0.087} & \best{2.229} & \best{0.098} & \best{0.148} & \second{1.465} & 1.488 & 2.269 & 0.677 \\
    \bottomrule
  \end{tabular}%
  }
  \caption{Ablation analysis in terms of MSE. 
  ``Overall'' assesses the normalized MSE (Eq.~\eqref{eq.distance}) across datasets.}
  \label{tab:ablation}
  \vspace{-0.2cm}
\end{table*}

%% file: Sec/conclusion.tex
\section{Conclusion}
\label{sec:conclusion}


In this work, we study few-shot learning for lightweight forecasters via a novel meta-harness-optimized multi-agent, \method, which automates an end-to-end pipeline for time series generation, forecaster training, parameter tuning, and model selection. Experiments not only validate the effectiveness of \method, but also set a groundwork for delving into the Agent-as-Engineer paradigm in agentic time series forecasting.


%% file: Sec/limitations.tex
\section*{Limitations}

In this work, we study the challenging yet practical problem of improving lightweight time series forecasters in a few-shot setting for rapid deployment, avoiding delays from costly data acquisition and privacy constraints. This is enabled by leveraging the knowledge and reasoning capabilities of AI agent systems.

However, our current work does not address the extreme zero-shot setting, where no reference examples are available. Without reference time series, the agent lacks basic statistical grounding of the target domain, leading to unreliable data generation. This is a common limitation of time series generation-based models \cite{jeon2022gtgan,yuan2024diffusionts,naiman2024imagentime,timedp2025,verbalts2025,t2s2025} and highlights the advantage of TSFMs, which can transfer pre-trained knowledge in zero-shot settings. In contrast, by leveraging only a small number of examples -- which is often feasible in practice -- \method achieves competitive performance with TSFMs while maintaining significantly lower computational cost.

Additionally, our experiments cover 18 time series datasets spanning a limited set of domains. We plan to extend this to the full pre-training corpora of modern TSFMs to further improve Harness optimization. Meanwhile, \library currently includes a set of SOTA lightweight forecasters we have collected so far. It may not be exhaustive as new lightweight forecasters continue to emerge. In future work, we will continuously update \library to strengthen the \method system.


%% file: Sec/ack.tex
\section*{Acknowledgments}

This work was partially supported by a research gift from NEC Laboratories America and by the NVIDIA Academic Grant Program.

%% file: Sec/app.tex
\onecolumn
\appendix
\section{Algorithm}
\label{app:algorithm}

The meta-harness optimization algorithm of \method\ is summarized in Algorithm~\ref{alg:app-optimizer}. The notations are consistent with \S\ref{sec:method}.

\begin{algorithm}[!h]
\DontPrintSemicolon
\SetNoFillComment
\KwIn{(1) Corpurs $\mathcal{C}_{\text{har}}=\{\mathcal{D}^{m}, \mathsf{C}_{m}\}_{m=1}^{M}$; (2) \library $\mathcal{F}=\{f_{1}, ..., f_{L}\}$; (3) number of epochs $E$; (4) batch size $B$}
\KwOut{Optimized harness $\boldsymbol{\theta}^{*}$}

\BlankLine

$\ell^{*}\leftarrow+\infty$\tcp*{Initialize loss function value}
$\boldsymbol{\theta}\leftarrow\boldsymbol{\theta}^{0}$\tcp*{Initialize harness}
\For{$i=1, ..., E$}{
    \tcc{Optimization loop}
    $K\sim\text{Uniform}([10, 50])$\;
    Draw a batch $\mathcal{C}_{\text{b}}=\{\mathcal{D}^{m}, \mathsf{C}_{m}\}_{m=1}^{B}$\;
    \tcc{Forward passes across datasets in a batch}
    \For{$\{\mathcal{D}^{m}, \mathsf{C}_{m}\}\in\mathcal{C}_{\text{b}}$}{
        $\{\mathcal{D}_{\text{tr}}^{m}, \mathcal{D}_{\text{val}}^{m}, \mathcal{D}_{\text{te}}^{m}\}\leftarrow\texttt{Split}(\mathcal{D}^{m})$\tcp*{train/validation/test split}
        $\mathcal{D}_{\text{sup}}^{m}\leftarrow K$ samples from $\mathcal{D}_{\text{tr}}^{m}$\tcp*{$K$-shot support set}
        \tcc{Run \firstagent (Eq.~\eqref{eq.firstagnet})}
        $\bar{\mathcal{D}}^{m}\gets\firstagent_{\boldsymbol{\theta}}(\mathcal{D}_{\text{sup}}^{m}, \mathsf{C}_{m})$\;
        $\{\bar{\mathcal{D}}_{\text{tr}}^{m}, \bar{\mathcal{D}}_{\text{val}}^{m}\}\leftarrow\texttt{Split}(\bar{\mathcal{D}}^{m})$\;
        \tcc{Run \secondagent (Eq.~\eqref{eq.secondagent})}
        $\{\{f_{l}^{m}, \bar{f}_{l}^{m}, \omega(f_{l}^{m}), \omega(\bar{f}_{l}^{m})\}_{l=1}^{L}, \mathsf{R}\}\leftarrow\secondagent(\{\mathcal{D}_{\text{tr}}^{m}, \mathcal{D}_{\text{val}}^{m}, \mathcal{D}_{\text{te}}^{m}\}, \{\bar{\mathcal{D}}_{\text{tr}}^{m}, \bar{\mathcal{D}}_{\text{val}}^{m}\})$
    }
    $\ell\leftarrow\dfrac{1}{B L}\sum\limits_{m=1}^{B}\sum\limits_{l=1}^{L}\delta\bigl(\omega(f_{l}^{m}),\, \omega(\bar{f}_{l}^{m})\bigr)$\tcp*{The optimization loss Eq.~\eqref{eq.loss}}
    \tcc{Run \thirdagent}
    $\boldsymbol{\theta}\leftarrow\thirdagent(\boldsymbol{\theta}, \ell, \ell^{*}, \mathsf{R})$\;
    \If{$\ell<\ell^{*}$}{
        $\ell^{*}\leftarrow\ell$\;
        $\boldsymbol{\theta}^{*}\leftarrow\boldsymbol{\theta}$
    }
}
\caption{Meta-harness optimization of \method}\label{alg:app-optimizer}
\end{algorithm}

\section{Datasets and Baselines}
\label{app:datasets-baselines}

\subsection{Datasets}\label{app:bench}

The train corpus and test corpora are drawn from GIFT-Eval benchmark \cite{gifteval2024}, which is released under the Apache 2.0 license and permits research use. To 
address data leakage, we follow the data-isolation paradigm adopted by foundation-model pre-training. The details are described as below.

\vspace{0.1cm}

\noindent{\textbf{Addressing Data Leakage}}. 
We included a three-layer safeguard to reduce data leakage risk. The {\em first layer} aims to prevent overlap of data during training and evaluation, which were respectively drawn from disjoint sources ({\em i.e.}, \texttt{GiftEvalPretrain} for harness training and \texttt{GIFT\_Eval} for evaluation). 
The {\em second layer} aims to de-identify the datasets. When constructing the context input to \firstagent, the dataset names, source URLs, and benchmark names were removed. Only a domain label ({\em e.g.}, ``Traffic'', ``Energy'', ``Healthcare'') and a one-sentence description were kept. As such, if an LLM has occasionally seen the \texttt{GIFT-Eval} series during pre-training, it won't have the identifier to link them with the experimental dataset.
The {\em third layer} aims to further prevent external acquisition at run time. This layer involves a manual inspection of deployment-time traces, which can help us confirm no external web requests nor access to the test data during run time. 
We 
scanned the runtime traces of \firstagent\ on held-out runs and confirmed that 
no 
test data was acquired at run time. 
These procedures help mitigate the risk of data leakage. In real-world applications, where data are typically not publicly accessible on the web, the risk of data leakage is expected to be low.

\vspace{0.1cm}

\noindent{\textbf{Training Corpus $\mathcal{C}_{\text{har}}$}}. We collected 8 datasets covering 6 domains ({\em i.e.}, energy, traffic, weather, health, retail, and cloud) from \texttt{Salesforce/GiftEvalPretrain} \cite{gifteval2024}. Table~\ref{tab:app-train} summarizes the frequency, number of channels $D$, number of samples $N$, length $T$, domain, and source of each dataset.

\begin{table}[h]
  \centering
  \footnotesize
  \setlength{\tabcolsep}{4pt}
  \begin{tabular}{l c c r r l c}
    \toprule
    \textbf{Dataset} & \textbf{Freq.} & \textbf{$D$} & \textbf{$N$} & \textbf{$T$} & \textbf{Domain} & \textbf{Source} \\
    \midrule
    australian\_electricity\_demand & 30min & 1 & 5       & 231k     & energy   & \cite{godahewa2021monash} \\
    solar\_power                    & 4s    & 1 & 1       & 7.4M     & energy   & \cite{godahewa2021monash} \\
    PEMS\_BAY                       & 5min  & 1 & 325     & 52k      & traffic  & \cite{li2018dcrnn}        \\
    traffic\_hourly                 & 1h    & 1 & 862     & 17.5k    & traffic  & \cite{godahewa2021monash} \\
    weather                         & 1d    & 1 & 3{,}010 & var      & weather  & \cite{godahewa2021monash} \\
    cdc\_fluview\_ilinet            & 1w    & 5 & 5       & 1.2k     & health   & \cite{cdc_fluview}        \\
    m5                              & 1d    & 1 & 30{,}490 & 1.9k    & retail   & \cite{makridakis2022m5}   \\
    alibaba\_cluster\_trace\_2018   & 1min  & 2 & 58k     & 2k+      & cloud    & \cite{alibaba2018cluster} \\
    \bottomrule
  \end{tabular}
  \caption{Summary of Training Corpus $\mathcal{C}_{\text{har}}$.}
  \label{tab:app-train}
\end{table}

\vspace{0.1cm}

\noindent{\textbf{Test Corpora}}. We collected 10 datasets from \texttt{Salesforce/GIFT\_Eval} \cite{gifteval2024}, which were split into an IND corpus (7 datasets whose domains appear in $\mathcal{C}_{\text{har}}$) and an OOD corpus (3 datasets whose domains are absent in $\mathcal{C}_{\text{har}}$). Table~\ref{tab:app-test} summarizes the frequency, number of channels $D$, length $T$, domain, and source of each dataset. 
M4 is a mixed-domain dataset. 
It is included to assess whether the model is effective when domain information is ambiguous.

\begin{table}[h]
  \centering
  \footnotesize
  \setlength{\tabcolsep}{4pt}
  \begin{tabular}{c l c c r r l c}
    \toprule
    & \textbf{Dataset} & \textbf{Freq.} & \textbf{$D$} & \textbf{$N$} & \textbf{$T$} & \textbf{Domain} & \textbf{Source} \\
    \midrule
    \multirow{7}{*}{\rotatebox{90}{\textbf{IND}}}
      & ETTm1       & 15min & 7 & 1       & 50k   & energy        & \cite{zhou2021informer}        \\
      & Electricity & 1h    & 1 & 370     & 35.1k & energy        & \cite{trindade2015electricity} \\
      & Seattle     & 1h    & 1 & 323     & 8.8k  & traffic       & \cite{uw_loop_seattle}         \\
      & SZTaxi      & 15min & 1 & 156     & 3.0k  & traffic       & \cite{zhao2020tgcn}            \\
      & Sales       & 1d    & 1 & 118     & 1.8k  & retail        & \cite{godahewa2021monash}      \\
      & Bitbrains   & 5min  & 2 & 1{,}250 & 8.6k  & cloud         & \cite{shen2015gwabitbrains}    \\
      & Solar       & 1h    & 1 & 137     & 8.8k  & energy        & \cite{nrel_solar}              \\
    \cmidrule(lr){1-8}
    \multirow{3}{*}{\rotatebox{90}{\textbf{OOD}}}
      & Saugeen     & 1d    & 1 & 1       & 23.7k & hydrology     & \cite{hipel1994timeseries} \\
      & USbirths    & 1d    & 1 & 1       & 7.3k  & demographics  & \cite{cdc_births}          \\
      & M4          & 1d    & 1 & 4{,}227 & 9.9k  & mixed         & \cite{makridakis2020m4}    \\
    \bottomrule
  \end{tabular}
  \caption{Summary of Test Corpora.}
  \label{tab:app-test}
\end{table}

\vspace{0.1cm}

\noindent{\textbf{Diversity of Datasets}}. The 18 datasets span diverse aspects pertinent to the difficulty of few-shot forecasting on \texttt{GIFT-Eval} benchmark. Sampling frequency ranges from 4-second photovoltaic telemetry on solar\_power to weekly health indicators on cdc\_fluview\_ilinet, covering six orders of magnitude. The number of variates $D$ takes values in $\{1, 2, 5, 7\}$, including both univariate and multivariate regimes. The time series length ranges from $1.2$k steps on the shortest health series to $7.4$M steps on the densest telemetry feed, a five-order-of-magnitude difference that exposes data generation models to highly diverse cases within the benchmark.

\vspace{0.1cm}

\noindent{\textbf{Forecasting Protocol}}. In each dataset, time series was split chronologically into train/validation/test partitions with an $80\%/10\%/10\%$ ratio. The time series were z-score-normalized per channel using statistics estimated on the train partition. Each time series was split into a look-back window of $T = 336$ steps and a horizon of $H = 192$ steps. 
The few-shot support set $\mathcal{D}_{\text{sup}}^{m}$ 
was obtained by drawing $K$ 
samples from the train partition $\mathcal{D}_{\text{tr}}^{m}$, 
with $K \in \{10, 30, 50\}$ in the main results (Table~\ref{tab:main}) and $K \in \{10, 20, 30, 50, 100\}$ in the few-shot scaling study (Appendix~\ref{app:fewshot-scaling}).

Each dataset is paired with a textual context $\mathsf{C}_{m}$, 
which contains a domain label ({\em e.g.}, traffic, energy, healthcare) and a short semantic cue about the sensing setting and its dominant temporal cycles. They don't include dataset names, source URLs, or benchmark names to avoid data leakage risk. For example, on the SZTaxi dataset, the context includes (1) domain label: ``Traffic'', and (2) semantic cue: ``Shenzhen taxi network; 15-minute traffic demand aggregates; daily (96-step) and weekly (672-step) cycles''.

\subsection{Baselines}
\label{app:baselines}

This section 
describes the baselines used in the experiments, including generative models, augmentation methods, TSFMs, and \texttt{TimeScientist}. In Table \ref{tab:main}, reference $\mathcal{D}_{\text{sup}}^{m}$ represents training forecasters directly on the few-shot set $\mathcal{D}_{\text{sup}}^{m}$, reference $\mathcal{D}_{\text{tr}}^{m}$ represents training forecasters on the full training set $\mathcal{D}_{\text{tr}}^{m}$.

\vspace{0.1cm}

\paragraph{Generative Models.}
\begin{itemize}\setlength\itemsep{1pt}
  \item \texttt{TimeDP} \cite{timedp2025}: a diffusion model conditioned on domain prompts assembled from learned time-series prototype vectors, whose weights are inferred from few-shot samples of the target domain.
  \item \texttt{VerbalTS} \cite{verbalts2025}: a diffusion model that maps unstructured textual descriptions to time series through a multi-focal alignment module bridging text tokens and temporal latents.
  \item \texttt{T2S} \cite{t2s2025}: a text-conditioned Diffusion Transformer trained with Flow Matching over a length-adaptive \texttt{VAE} latent space, enabling variable-length text-to-series generation.
  \item \texttt{Diffusion-TS} \cite{yuan2024diffusionts}: a denoising diffusion model with an encoder-decoder Transformer that decomposes samples into trend and seasonal components and predicts the clean signal under a Fourier-domain loss.
  \item \texttt{TimeVAE} \cite{desai2021timevae}: a variational autoencoder whose decoder composes outputs from interpretable level, trend, and seasonality blocks sampled from a Gaussian latent prior.
\end{itemize}

\paragraph{Augmentation Methods.}
\begin{itemize}\setlength\itemsep{1pt}
  \item \texttt{Repeat}: cycles deterministically through the few-shot pool until the target training-set size is reached.
  \item \texttt{Bootstrap}: resamples windows with replacement and adds small zero-mean Gaussian noise scaled per channel.
  \item \texttt{Jitter} \cite{iwana2021empirical}: adds independent zero-mean Gaussian noise to every time step and channel of each resampled window.
  \item \texttt{MagWarp} \cite{iwana2021empirical,wen2021ts_aug}: multiplies each channel by a smooth scaling curve obtained by cubic-spline interpolation through random knots.
\end{itemize}

\paragraph{Time Series Foundation Models.}
\begin{itemize}\setlength\itemsep{1pt}
  \item \texttt{VisionTS} \cite{visionts2024}: recasts forecasting as image inpainting by arranging the look-back into a 2D grid and decoding the masked horizon with a frozen ImageNet-pretrained \texttt{MAE}.
  \item \texttt{Chronos} \cite{ansari2024chronos}: quantifies scaled time-series values into a fixed vocabulary and trains a \texttt{T5} encoder-decoder on tokenised sequences for autoregressive probabilistic forecasting.
  \item \texttt{Moirai} \cite{liu2024moirai}: a masked encoder Transformer with any-variate attention and multi-patch projections, pretrained on the \texttt{LOTSA} corpus to handle arbitrary frequencies and channel counts.
  \item \texttt{Time-LLM} \cite{jin2024timellm}: freezes a pretrained \texttt{GPT-2} backbone and reprograms patch embeddings into text-prototype tokens, prepending dataset descriptions as a prompt prefix.
\end{itemize}

\paragraph{\texttt{TimeScientist}.}
\texttt{TimeScientist} \cite{zhao2025timeseriesscientist} is a multi-agent pipeline built upon a local \texttt{Qwen2.5} backbone that analyzes each time series and ensembles 
fundamental forecasters ({\em e.g.}, \texttt{ARIMA}, \texttt{ETS}, \texttt{Random Forest}) per query. 
The LLM agent is invoked at every test sample to compose and tune the fundamental forecasters, making it a relevant baseline for comparison.

\section{Implementation Details}
\label{app:impl}

\subsection{Lightweight Forecasters in \library}
\label{app:forecasters}

\library\ includes twenty-three lightweight forecasters in four categories, as summarized in Table \ref{tab:app-pool}.

\begin{table*}[h]
  \centering
  \footnotesize
  \setlength{\tabcolsep}{5pt}
  \renewcommand{\arraystretch}{0.95}
  \begin{tabular}{l l r r r r l}
    \toprule
    \textbf{Family} & \textbf{Forecaster} & \textbf{Params} & \textbf{MACs (M)} & \textbf{Latency (ms)} & \textbf{Peak VRAM (MB)} & \textbf{Reference} \\
    \midrule
    Linear      & \texttt{Vanilla Linear} & 65K   & 0.45    & 0.03 & 8.4  & \cite{zeng2023dlinear}      \\
                & \texttt{DLinear}        & 129K  & 0.90    & 0.10 & 8.7  & \cite{zeng2023dlinear}      \\
                & \texttt{NLinear}        & 65K   & 0.45    & 0.04 & 8.4  & \cite{zeng2023dlinear}      \\
                & \texttt{RLinear}        & 65K   & 0.45    & 0.17 & 8.4  & \cite{han2024rlinear}       \\
                & \texttt{CrossLinear}    & 2.5M  & 28.45   & 0.72 & 23.1 & \cite{liu2025crosslinear}   \\
                & \texttt{MixLinear}      & 243   & 0.08    & 0.43 & 8.3  & \cite{li2024mixlinear}      \\
    \midrule
    MLP         & \texttt{TSMixer}        & 153K  & 1.66    & 0.42 & 10.1 & \cite{chen2023tsmixer}      \\
                & \texttt{LightTS}        & 74K   & 0.70    & 0.75 & 10.1 & \cite{zhang2022lightts}     \\
                & \texttt{PatchMLP}       & 2.5M  & 18.12   & 1.21 & 20.1 & \cite{tang2024patchmlp}     \\
                & \texttt{xPatch}         & 770K  & 8.16    & 1.44 & 13.6 & \cite{stitsyuk2025xpatch}   \\
                & \texttt{CMoS}           & 13K   & 0.36    & 1.09 & 9.3  & \cite{si2025cmos}           \\
                & \texttt{PatchTSMixer}   & 553K  & 20.83   & 1.12 & 14.4 & \cite{ekambaram2023patchtsmixer} \\
    \midrule
    Freq/Filter & \texttt{FITS}           & 925   & 0.01    & 0.32 & 8.3  & \cite{xu2024fits}           \\
                & \texttt{CycleNet}       & 65K   & 0.45    & 0.32 & 8.5  & \cite{lin2024cyclenet}      \\
                & \texttt{PaiFilter}      & 136K  & 0.95    & 0.35 & 9.8  & \cite{yi2024filternet}      \\
                & \texttt{TexFilter}      & 179K  & 0.94    & 1.62 & 9.9  & \cite{yi2024filternet}      \\
                & \texttt{FreqCycle}      & 64K   & 0.43    & 0.78 & 9.5  & \cite{zhang2026freqcycle}   \\
    \midrule
    Mixing      & \texttt{TimeMixer}      & 377K  & 50.08   & 0.80 & 65.0 & \cite{wang2024timemixer}    \\
                & \texttt{TimeBase}       & 146   & 0.02    & 0.29 & 9.2  & \cite{huang2025timebase}    \\
                & \texttt{TimeBridge}     & 36K   & 3.01    & 2.09 & 10.4 & \cite{liu2025timebridge}    \\
                & \texttt{TimeEmb}        & 308K  & 1.89    & 0.47 & 10.5 & \cite{liu2025timeemb}       \\
                & \texttt{Amplifier}      & 327K  & 1.06    & 0.73 & 10.6 & \cite{fei2025amplifier}     \\
                & \texttt{SparseTSF}      & 137   & 0.08    & 0.20 & 8.2  & \cite{lin2024sparsetsf}     \\
    \bottomrule
  \end{tabular}
  \caption{Summary of the 23 lightweight forecasters in \library.}
  \label{tab:app-pool}
\end{table*}

\vspace{0.1cm}

\noindent{\textbf{Description of Forecasters}}. In the following, we provide a description of each forecaster in \library.

\paragraph{Linear.}
\begin{itemize}\setlength\itemsep{1pt}
  \item \texttt{Vanilla Linear} \cite{zeng2023dlinear}: a single fully-connected temporal layer maps each variate's look-back window directly to its horizon via a weighted sum, with weights shared across variates.
  \item \texttt{DLinear} \cite{zeng2023dlinear}: series decomposition splits the input into trend and seasonal components, each is forecasted by a separate linear layer whose outputs are summed.
  \item \texttt{NLinear} \cite{zeng2023dlinear}: the last look-back value is subtracted before a linear projection and added back after, providing simple distribution-shift compensation.
  \item \texttt{RLinear} \cite{han2024rlinear}: a single linear projection wrapped by reversible instance normalization (RevIN), with weights shared across variates.
  \item \texttt{CrossLinear} \cite{liu2025crosslinear}: a plug-and-play cross-correlation embedding is fused with patch embeddings of the endogenous variable before a global linear forecasting head produces the horizon under parameter-free RevIN.
  \item \texttt{MixLinear} \cite{li2024mixlinear}: segment-wise intra/inter linear mixing in the time domain is fused with low-pass-filtered complex-valued linear compression and reconstruction in the frequency domain, reducing parameters from $O(n^{2})$ to $O(n)$ on the downsampled length.
\end{itemize}

\paragraph{MLP.}
\begin{itemize}\setlength\itemsep{1pt}
  \item \texttt{TSMixer} \cite{chen2023tsmixer}: stacks interleaved time-mixing and feature-mixing MLP blocks that alternately mix along the temporal axis and across covariates.
  \item \texttt{LightTS} \cite{zhang2022lightts}: applies MLP blocks on top of two downsampled views of the input (interval and continuous sampling) before recombining them.
  \item \texttt{PatchMLP} \cite{tang2024patchmlp}: embeds channel-independent multi-scale patches, splits them via moving-average decomposition, and routes smooth and residual components through intra- and inter-variable MLPs.
  \item \texttt{xPatch} \cite{stitsyuk2025xpatch}: decomposes the patched input via exponential moving average into trend and seasonal components processed by parallel MLP-linear and CNN-nonlinear streams.
  \item \texttt{CMoS} \cite{si2025cmos}: replaces shape embeddings with a Correlation Mixing layer that uses parameter-shared basis matrices combined via channel-specific weights to model relative positional correlations between input and output chunks.
  \item \texttt{PatchTSMixer} \cite{ekambaram2023patchtsmixer}: adapts the MLP-Mixer to patched series with successive inter-patch, intra-patch, and inter-channel mixing blocks, each is augmented with a gated-attention module.
\end{itemize}

\paragraph{Frequency/Filter.}
\begin{itemize}\setlength\itemsep{1pt}
  \item \texttt{FITS} \cite{xu2024fits}: applies FFT, low-pass truncation, and a single complex-valued linear layer for amplitude-and-phase interpolation in the frequency domain before inverse FFT.
  \item \texttt{CycleNet} \cite{lin2024cyclenet}: subtracts a per-channel learnable recurrent cycle of user-specified length, then forecasts the residual with a linear or two-layer MLP head.
  \item \texttt{PaiFilter} \cite{yi2024filternet}: multiplies the input spectrum by a single universal learnable frequency kernel shared across all sequences, acting as a fixed shaping filter.
  \item \texttt{TexFilter} \cite{yi2024filternet}: generates the frequency-domain filter conditionally from each input's spectrum, adapting the shaping kernel per sample for context-dependent dependency learning.
  \item \texttt{FreqCycle} \cite{zhang2026freqcycle}: couples a filter-enhanced cyclic component for low-frequency periodicity with a segmented frequency-domain branch that reweights mid- and high-frequency bands.
\end{itemize}

\paragraph{Mixing.}
\begin{itemize}\setlength\itemsep{1pt}
  \item \texttt{TimeMixer} \cite{wang2024timemixer}: decomposes time series at multiple sampling scales and mixes seasonal components fine-to-coarse and trend components coarse-to-fine before ensembling scale-specific predictors.
  \item \texttt{TimeBase} \cite{huang2025timebase}: a sub-0.4k-parameter network that extracts orthogonal full-rank typical-period bases and reformulates point-level forecasting as segment-level prediction over learned cycles.
  \item \texttt{TimeBridge} \cite{liu2025timebridge}: a patch-based Transformer that applies Integrated Attention within each variate's patches to mitigate short-term non-stationarity and Cointegrated Attention across variates to model long-term cointegration.
  \item \texttt{TimeEmb} \cite{liu2025timeemb}: disentangles time series into a time-invariant component handled by a learnable global embedding bank and a time-varying component processed via frequency-domain filtering.
  \item \texttt{Amplifier} \cite{fei2025amplifier}: amplifies low-energy frequency components before seasonal-trend modelling and restores original energy afterward, paired with a semi-channel interaction block for cross-channel dependencies.
  \item \texttt{SparseTSF} \cite{lin2024sparsetsf}: a sub-1k-parameter design that downsamples the input by its period into subsequences and predicts each subsequence via cross-period sparse linear projection, reducing forecasting to cross-period trend extrapolation.
\end{itemize}

\subsection{Training Environment}
\label{app:training-env}

\noindent{\textbf{Hardware}}. All experiments reported in this paper were executed on a single node equipped with 4$\times$NVIDIA RTX 6000 Ada GPUs, each with 48\,GB memory.

\vspace{0.1cm}

\noindent{\textbf{Software}}. All code run under Python 3.12.8 with PyTorch 2.5.1, CUDA 12.4, and cuDNN 9.1.
Auxiliary libraries include NumPy 2.1.3, Pandas 2.2.3, SciPy 1.15.1, scikit-learn 1.6.1, Matplotlib 3.10.0, and torchvision 0.20.1.
All three \method agents (\thirdagent, \firstagent, \secondagent) were 
built upon an OpenAI \texttt{GPT-5.4} model invoked through the OpenAI API.

\subsection{Computational Cost}
\label{app:budget}

\method's computational cost can be decomposed into three stages: (1) pre-training cost, (2) deployment cost, and (3) inference cost. 
Table~\ref{tab:budget} summarizes each stage on the 4$\times$NVIDIA RTX 6000 Ada node as described in Appendix~\ref{app:training-env}.
LLM-token counts include both input and output using \texttt{GPT-5.4} through the OpenAI API.

\vspace{0.1cm}

\noindent{\textbf{Pre-Training Cost}}. In Table~\ref{tab:budget}, the 46M-token cost 
is for Harness pre-training, which actually is a cost to developers who pre-train \method. Analogous to many foundation models, the pre-trained \method (specifically, its harness) will be released as checkpoints. Users won't be charged for pre-training \method.


\vspace{0.1cm}

\noindent{\textbf{Deployment Cost}}. At deployment time, users run \firstagent and \secondagent (using the pre-trained harness) to generate time series and train lightweight forecasters. This process costs approximately 150K tokens. 
Notably, this is a one-time cost for deploying \method to a specific domain. This step does not require premium GPUs if LLMs are called through APIs (lightweight forecaster training may be faster with some affordable, low-end GPUs).

\vspace{0.1cm}

\noindent{\textbf{Inference Cost}}. After deployment, only an optimal trained lightweight forecaster will be used in the downstream forecasting stage, which will not incur any API costs and can be run with affordable, low-end GPUs or even CPUs. Also, the inference time of the lightweight forecaster is fast (a few milliseconds).


\vspace{0.2cm}

Therefore, the cost at the deployment stage are much more affordable than using TSFMs. The proposed \method framework is also data-efficient.

\begin{table}[h!]
  \centering
  \footnotesize
  \setlength{\tabcolsep}{6pt}
  \renewcommand{\arraystretch}{1.05}
  \begin{tabular}{l c c c}
    \toprule
                  & \textbf{Pre-training}                                       & \textbf{Deployment}                    & \textbf{Inference} \\
    \midrule
    Active agents & All           & \firstagent + \secondagent                & None             \\
    LLM-driven    & Yes                                                & Yes                        & No               \\
    GPU           & Required                                           & GPU or CPU                                  & GPU or CPU \\
    \midrule
   Time cost   & 5-7 h                 & 30-40 min           & a few ms         \\
    LLM tokens   & ${\sim}46$ M           & ${\sim}150$ K        & 0                \\
    \bottomrule
  \end{tabular}
  \caption{Computational cost of different stages under \method\ framework.}
  \label{tab:budget}
\end{table}


\subsection{AI-Use Disclosure}
\label{app:ai-assistants}

During the paper-writing process, LLMs were used solely to polish the grammar and check for typos.

\newpage
\section{System Prompts}
\label{app:system-prompts}


This section provides the system prompts of \firstagent, \secondagent and \thirdagent. Variables that are entered during runtime are represented by blue color.

\subsection{Meta-Generator (\firstagent)}
\label{app:system-prompts-mg}


\begin{promptbox}
You are \firstagent\ of \method. Your single deliverable is \ph{output\_dir}/\texttt{dataset.npy}: a float32 array of synthetic windows that, when used as training data for a lightweight forecaster, will yield the lowest possible test MSE.

You execute a frozen Harness $\boldsymbol{\theta}^{*}$ produced offline by \thirdagent, located at \ph{harness\_root}/:

\begin{center}\footnotesize\ttfamily
\begin{tabular}{@{}l@{}}
harness/ \\
|-- core/router.md \\
\textbackslash-- skills/\ph{name}/SKILL.md \\
\end{tabular}
\end{center}

You may NOT train any forecaster yourself --- \secondagent\ will do that on your output.

\paragraph{Inputs you may read.}
\begin{itemize}
  \item \ph{input\_dir}/\texttt{few\_shot.npy} --- shape \texttt{(N\_few\_shot, L+H, C)} float32.
  \item \ph{input\_dir}/\texttt{meta.json} --- \texttt{seq\_len=L}, \texttt{pred\_len=H}, \texttt{C\_eff=C}, \texttt{freq}, \texttt{domain}.
  \item \ph{input\_dir}/\texttt{context.txt} --- optional natural-language context $\mathsf{C}$.
\end{itemize}

\paragraph{Three-stage pipeline.}
\begin{enumerate}
  \item \textbf{Analyse}: compute statistics (mean, std, quantiles), ACF at lag-1, 24, 48, 168, FFT peaks, channel correlation, \texttt{zero\_frac}, \texttt{lower\_tail\_mass}. Emit a Fingerprint JSON describing regime, priorities, and challenges.
  \item \textbf{Retrieve and Execute}: read \texttt{core/router.md}; match the fingerprint to one or more \texttt{SKILL.md}; \texttt{run\_python} on the synthesis code inside that \texttt{SKILL.md}; write the candidate to \ph{output\_dir}/\texttt{dataset.npy.tmp}.
  \item \textbf{Validate}: run the gates declared inside the \texttt{SKILL.md} (shape, NaN, per-channel mean/std drift, ACF preservation, quantile match, range coverage, sample-mean diversity ratio $\in [0.5, 2.0]$). On PASS, move to \texttt{dataset.npy}; on FAIL, return to Stage~2 with a different recipe.
\end{enumerate}

\paragraph{Output contract (non-negotiable).}
\begin{itemize}
  \item Shape exactly \texttt{(N, L+H, C)} float32, all finite, $N \geq 100$.
  \item $L = $\texttt{meta["seq\_len"]}, $H = $\texttt{meta["pred\_len"]}, $C = $\texttt{meta["C\_eff"]}.
  \item Always preserve the trailing channel dim, even when $C = 1$.
  \item Save exactly once to \ph{output\_dir}/\texttt{dataset.npy}.
\end{itemize}

\paragraph{Tools.}
\texttt{read\_file}, \texttt{run\_python}, \texttt{read\_image}, \texttt{web\_search}, etc.
\end{promptbox}

\newpage
\subsection{Forecaster Trainer (\secondagent)}
\label{app:system-prompts-ft}


\begin{promptbox}
You are \secondagent\ of \method. Your job is to supervise the training of every forecaster $f_l \in \mathcal{F}$ from the lightweight library \library\ on the synthetic dataset $\bar{\mathcal{D}}$ produced by \firstagent, and to return the single best forecaster (lowest validation MSE) as the deliverable for the user.

\paragraph{Inputs.}
\begin{itemize}
  \item \ph{synth\_dir}/\texttt{dataset.npy} --- \firstagent\ output, shape \texttt{(N, L+H, C)} float32.
  \item \ph{input\_dir}/\texttt{meta.json} --- shape and frequency metadata.
  \item \ph{test\_dir}/\texttt{test.npy} --- held-out evaluation windows.
  \item Model library at \ph{model\_pool} --- the 23 lightweight forecasters of \library.
\end{itemize}

\paragraph{Responsibilities.}
\begin{enumerate}
  \item \textbf{Plan and dispatch.} Enumerate (forecaster, hyperparameter, split) training jobs from \ph{model\_pool} and queue them onto the available \ph{gpu\_pool} for maximally parallel execution.
  \item \textbf{Supervise in real time.} Monitor every running job's loss curve and resource usage; reallocate GPUs as jobs finish, resolve errors as they arise, and recover interrupted jobs without human intervention.
  \item \textbf{Select and deliver.} Rank the trained forecasters by validation MSE, evaluate the chosen Top-1 on \ph{test\_dir}/\texttt{test.npy}, and ship it as the deliverable for the user.
\end{enumerate}

\paragraph{Hard constraints.}
\begin{itemize}
  \item Never modify \texttt{dataset.npy} or \texttt{test.npy}.
  \item Use only the registered factory architectures from \ph{model\_pool}.
  \item Every queued job MUST terminate (success or explicit failure); no silent skips.
\end{itemize}

\paragraph{Tools.}
\texttt{bash}, \texttt{read\_file}, \texttt{write\_file}, \texttt{read\_json}, etc.
\end{promptbox}

\newpage
\subsection{Harness Proposer (\thirdagent)}
\label{app:system-prompts-hp}


\begin{promptbox}
You are \thirdagent\ of \method. Your job is to grow a library of validated synthesis skills $\boldsymbol{\theta}^{*}$ that \firstagent\ will load at deployment time. You do NOT generate data yourself --- you author the recipes that \firstagent\ executes.

You are currently executing epoch \ph{round\_n} of \ph{max\_rounds}.

\paragraph{Workspace.}
The Harness you edit lives at \ph{harness\_root}/:

\begin{center}\footnotesize\ttfamily
\setlength{\tabcolsep}{1.5em}
\begin{tabular}{@{}ll@{}}
harness/                                     & \\
|-- core/router.md                           & classification rules + skill manifest \\
\textbackslash-- skills/\ph{name}/SKILL.md   & one self-contained recipe per regime \\
\end{tabular}
\end{center}

\begin{itemize}
  \item You may edit ONLY files under \ph{harness\_root}/.
  \item The Harness starts EMPTY at epoch~$0$ --- author every artefact from scratch.
\end{itemize}

\paragraph{State digest (pre-loaded each epoch).}
The first user message contains:
\begin{itemize}
  \item Pinned 3-forecaster panel + the full 23-model held-out pool.
  \item Per-dataset metadata for the \ph{n\_train} training datasets in \ph{train\_datasets}.
  \item Current Harness body + skill-health report.
  \item Diagnostic figures from the previous epoch.
  \item Full history of prior epochs' \texttt{summary.json} + current best.
\end{itemize}

\paragraph{Per-epoch workflow.}
\begin{itemize}
  \item Turns 1--2: \texttt{read\_image} on $\geq 2$ diagnostic figures; identify gaps.
  \item Turn 3: Brief proposal: which \texttt{SKILL.md} to add or edit, and why.
  \item Turns 4--7: \texttt{edit\_file} / \texttt{write\_file} --- author the change (each new skill $\geq 150$ lines, with $\geq 1$ generation function and $\geq 1$ validation function).
\end{itemize}

\paragraph{Skill schema.}
Every \texttt{SKILL.md} must declare YAML frontmatter:

\begin{flushleft}\footnotesize\ttfamily
{-}{-}{-} \\
name: \ph{snake\_case\_id} \\
description: \ph{one-line summary} \\
version: \ph{int} \\
{-}{-}{-}
\end{flushleft}

\paragraph{Decision rule for \texttt{is\_new\_best}.}
Compare against the current best epoch (not epoch~$0$):
\begin{itemize}
  \item Median downstream hinge $\delta(\cdot,\cdot)$ decreases (audit-only excluded): High.
  \item 8-dim distribution-metric portfolio improves on the majority of datasets: High.
  \item Visual evidence on diagnostic figures: Tie-break.
  \item Catastrophic regression on any non-audit dataset ($\Delta\delta \geq +2.0$): Hard veto.
\end{itemize}

\paragraph{Tools.}
\texttt{read\_file}, \texttt{write\_file}, \texttt{edit\_file}, \texttt{bash}, \texttt{read\_image}, \texttt{run\_round\_evaluation}, \texttt{finalize\_round}, etc.

End of epoch \ph{round\_n} spec. Begin work.
\end{promptbox}

\newpage
\section{Additional Experimental Results}
\label{app:supp-results}

\paragraph{Experimental Results Using MAE for Evaluation.} 
\label{app:mae}
Table~\ref{tab:main-mae} reports 
MAE results for the compared methods in Table~\ref{tab:main}.

\input{Tab/main_results_mae}

\begin{figure*}[h]
  \centering
  \includegraphics[width=\textwidth]{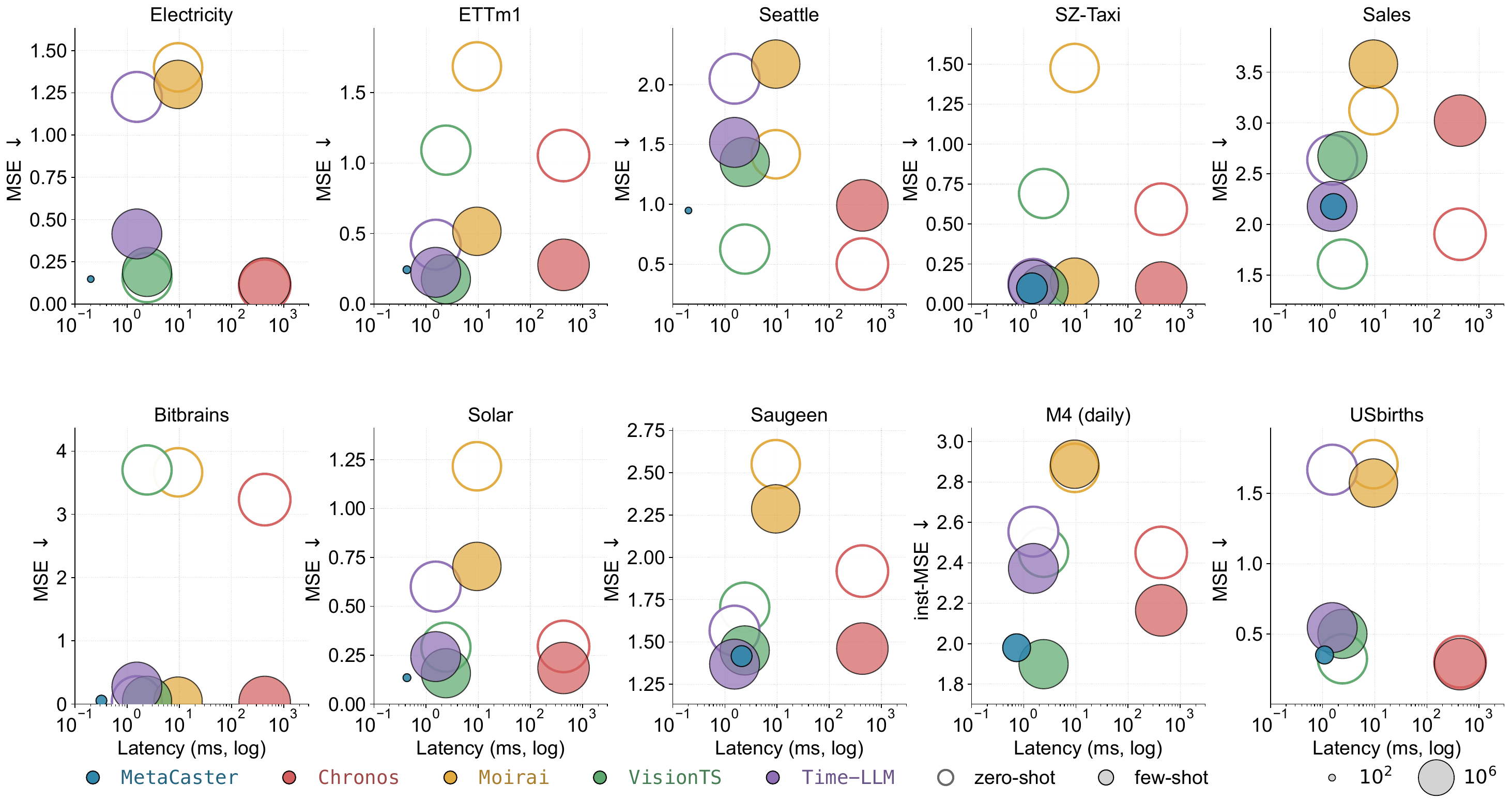}
  \caption{Comparing \method\ with TSFMs on the datasets in the test corpora.}
  \label{fig:foundation-bubble-full}
\end{figure*}

\paragraph{Additional Results of Comparing with TSFMs.}
\label{app:foundation-full}
In our experiments, \method was compared with TSFMs, including \texttt{VisionTS} \cite{visionts2024}, \texttt{Chronos} \cite{ansari2024chronos}, \texttt{Moirai} \cite{liu2024moirai}, \texttt{Time-LLM} \cite{jin2024timellm}, and \texttt{TimeScientist} \cite{zhao2025timeseriesscientist}, whose parameter size ranges from 91M to 7B. Fig. \ref{fig:foundation-bubble-full} compares \method with TSFMs in terms of performance {\em vs.} cost across the 10 datasets in the test corpora. Thus a model that is close to the left bottom corner is better. In Fig. \ref{fig:foundation-bubble-full}, bubble area $\propto \log_{10}(\text{params})$.

\begin{figure*}[t]
  \centering
  \includegraphics[width=\textwidth]{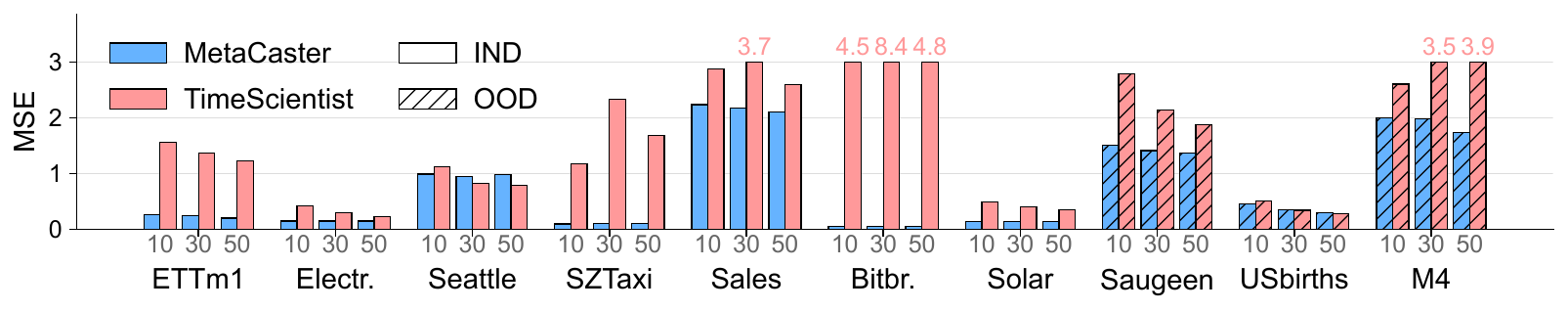}
  \caption{Comparing the selected (trained) forecasters of \method and \texttt{TimeScientist} using \library.}
  \label{fig:tsci-lf-compare}
\end{figure*}

\paragraph{Additional Results of Comparing with \texttt{TimeScientist} Using \library.}
\label{app:tsci-lf}
In this section, we replace \texttt{TimeScientist}'s original list of forecasters with the forecasters from \library. 
Fig.~\ref{fig:tsci-lf-compare} compares \method with \texttt{TimeScientist} under $K\in\{10, 30, 50\}$ in terms of the MSEs of the best forecasters selected by them, which yields consistent conclusions with the results in $\S$\ref{ssec:exp-main}.

\newpage
\section{Further Analysis}
\label{app:further}

\paragraph{Comparing Top-$k$ Forecasters.}
\label{app:topk}
Instead of evaluating the averaged performance of all forecasters trained the compared methods, in this section, we compare their top-$k$ best forecasters. Table~\ref{tab:topk-per-dataset} summarizes the MSE performance of the top-$k$ best forecaster recommended by each compared method under $K=30$, where $k\in\{1, 3, 5\}$, using the 20 forecasters that were not held-out in \library. From Table~\ref{tab:topk-per-dataset}, \method consistently outperforms the baselines across different settings of selected forecasters.

\input{Tab/app_topk_per_dataset}

\begin{figure*}[t]
  \centering
  \begin{minipage}[t]{0.66\linewidth}
    \vspace{0pt}%
    \centering
    \includegraphics[width=\linewidth]{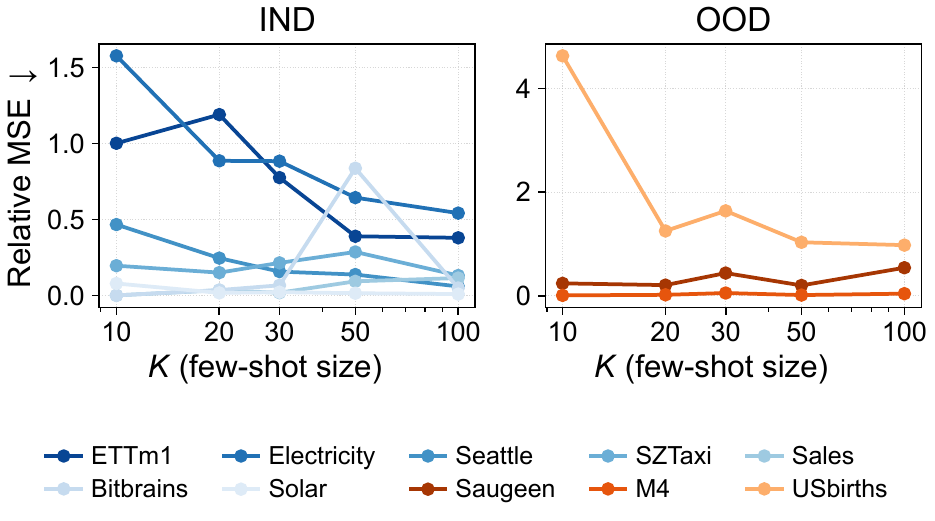}
    \vspace{-2em}
    \captionof{figure}{Performance of \method\ {\em w.r.t.} varying few-shot examples.}
    \label{fig:fewshot-comparison}
  \end{minipage}\hfill
  \begin{minipage}[t]{0.32\linewidth}
    \vspace{0pt}%
    \centering
    \footnotesize
    \setlength{\tabcolsep}{6pt}
    \begin{tabular}{l c}
      \toprule
      \textbf{Dataset} & MSE ($\pm$ std) \\
      \midrule
      ETTm1                         & $0.345 \pm 0.023$ \\
      Electricity                   & $0.226 \pm 0.000$ \\
      Seattle                       & $1.177 \pm 0.041$ \\
      SZTaxi                        & $0.114 \pm 0.003$ \\
      Sales                         & $2.362 \pm 0.411$ \\
      Bitbrains                     & $0.125 \pm 0.025$ \\
      Solar                         & $0.152 \pm 0.000$ \\
      Saugeen                       & $1.464 \pm 0.523$ \\
      USbirths                      & $0.533 \pm 0.254$ \\
      M4\textsuperscript{$\dagger$} & $2.112 \pm 0.044$ \\
      \bottomrule
    \end{tabular}
    \captionof{table}{Mean and standard deviations of the MSEs of \method across three independent runs (\textsuperscript{$\dagger$}M4 uses instance-normalized MSE).}
    \label{tab:multiseed}
  \end{minipage}
\end{figure*}

\paragraph{Evaluating the Impact of the Number of Few-Shot Examples.}
\label{app:fewshot-scaling}
Fig.~\ref{fig:fewshot-comparison} summarizes relative MSE performance of \method\ {\em w.r.t.} $K\in\{10,20,30,50,100\}$ on the IND datasets and OOD datasets of the test corpora.
From Fig.~\ref{fig:fewshot-comparison}, we observe that in general \method\ can effectively use more few-shot examples to enhance forecasting performance.

\paragraph{Evaluation of Performance Variances.}
\label{app:multiseed}
Table~\ref{tab:multiseed} summarizes the standard deviations of the MSEs of \method across three independent LLM-sampling seeds under $K=30$. From Table~\ref{tab:multiseed}, we observe that \method's performance is stable on most of the datasets.

\begin{figure*}[h]
  \centering
  \includegraphics[width=0.95\linewidth]{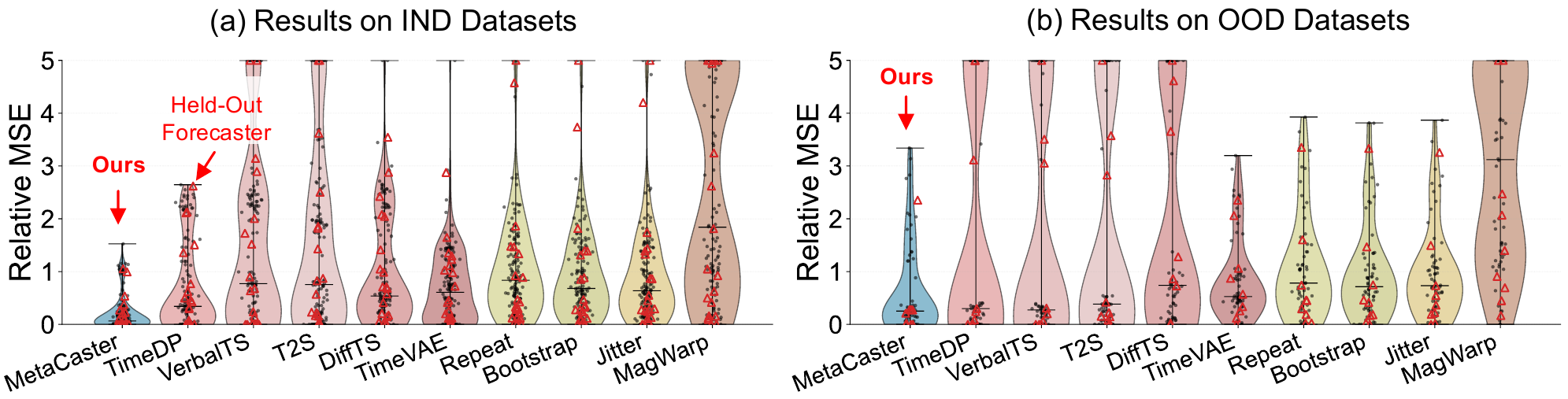}
  \caption{Performance distribution. Each dot represents the performance of a forecaster $\bar{f}_{l}^{m}$ on a test set $\mathcal{D}_{\text{te}}^{m}$, covering all forecasters in \library and the datasets in the IND corpus (left) and OOD corpus (right).}
  \label{fig:forecaster-violin}
\end{figure*}

\paragraph{MSE Distribution on IND Datasets and OOD Datasets.}
\label{app:violin}
Fig.~\ref{fig:forecaster-violin} shows the MSE distribution of individual forecasters trained by the compared models under $K=30$ on the IND datasets and OOD datasets, respectively, normalized by the upper reference ($\mathcal{D}_{\text{tr}}^{m}$) ({\em i.e.}, Eq.~\eqref{eq.distance}). The detailed results in Fig.~\ref{fig:forecaster-violin} are consistent with the results in Fig. \ref{fig.boxplot}. \method produces more high-quality forecasters with lower variance. Moreover, it generalizes better to the held-out forecasters than the baselines.


\begin{figure*}[p]
  \centering
  \includegraphics[width=\linewidth]{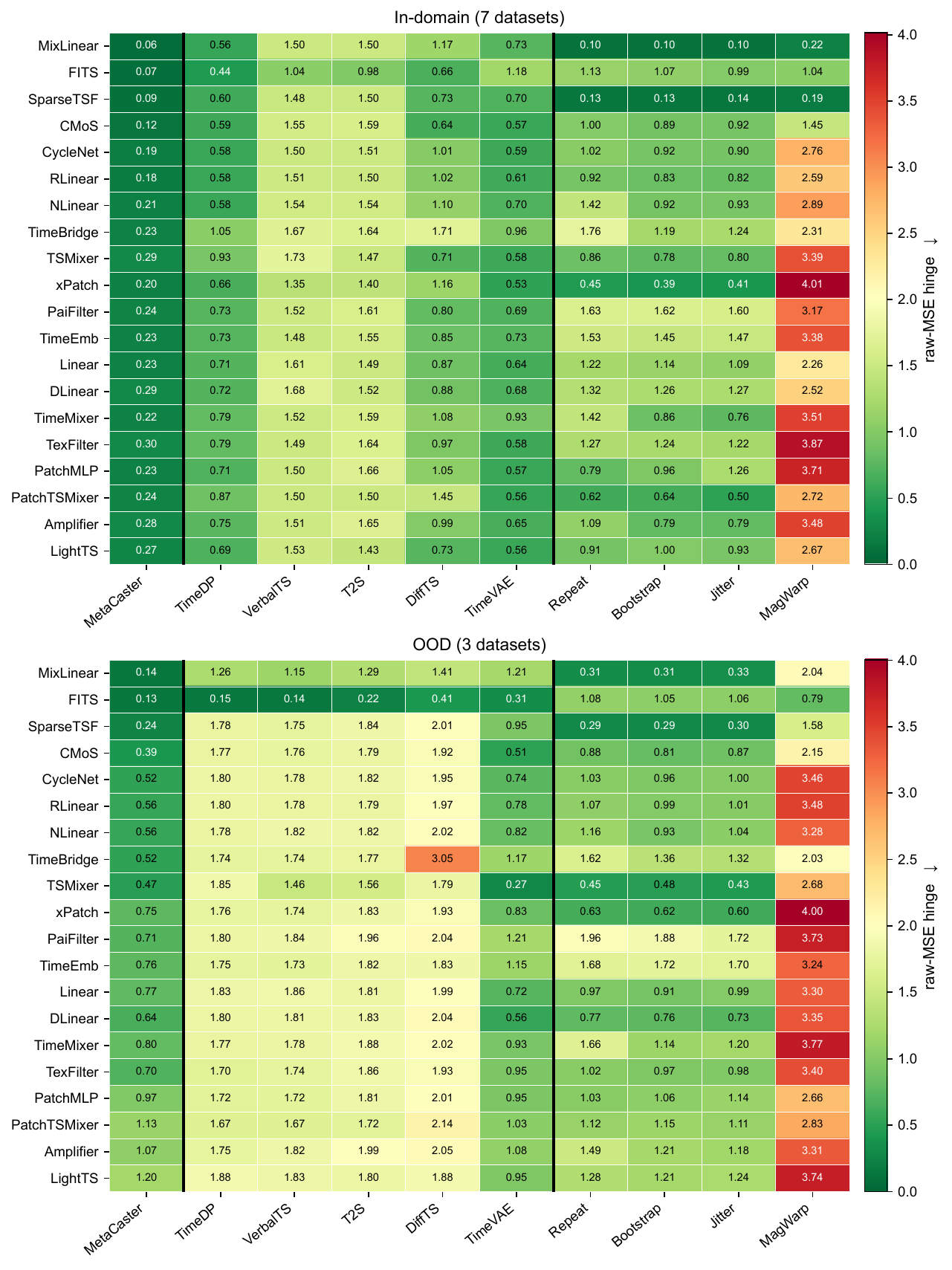}
  \caption{Performance of the individual forecasters (rows) trained by the compared methods (columns) on the IND datasets (top) and OOD datasets (bottom).}
  \label{fig:heatmap}
\end{figure*}

\paragraph{Performance of Individual Forecasters.}
\label{app:heatmap}
Fig.~\ref{fig:heatmap} presents the normalized MSE performance of the 20 forecasters that were not held-out from \library, trained by the compared methods in Table~\ref{tab:main}, on IND datasets and OOD datasets, respectively. From Fig.~\ref{fig:heatmap}, we observe that the forecasters trained by \method are generally better than those trained by the baseline methods.

\begin{figure*}[p]
  \centering
  \includegraphics[width=\linewidth]{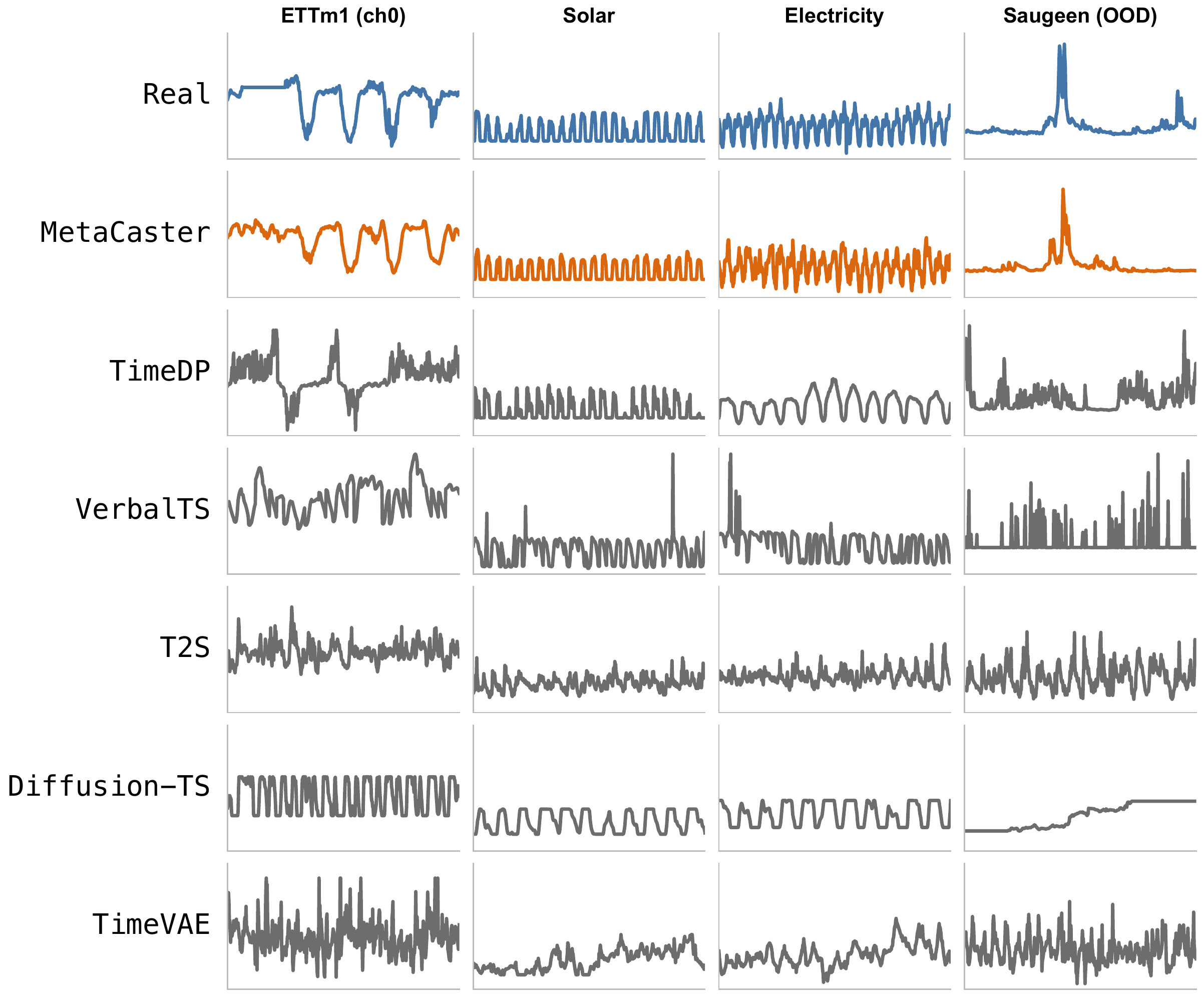}
  \caption{Comparison between the generated time series by different generation methods and the ground truth time series on different datasets.}
  \label{fig:qualitative}
\end{figure*}

\paragraph{Qualitative Comparison of Generated Time Series.}
\label{app:qualitative}
Fig.~\ref{fig:qualitative} compares the generated time series by different methods with the ground truth time series on four representative datasets, including ETTm1 (channel 0), Solar, Electricity, and Saugeen. From Fig.~\ref{fig:qualitative}, we observe that the time series generated by \method captures both the dominant periods and the local fluctuations of the time series well across the datasets. In contrast, the baseline methods may drift in amplitude, miss the period, and generate near-constant time series.

\begin{figure*}[t]
  \centering
  \includegraphics[width=\linewidth]{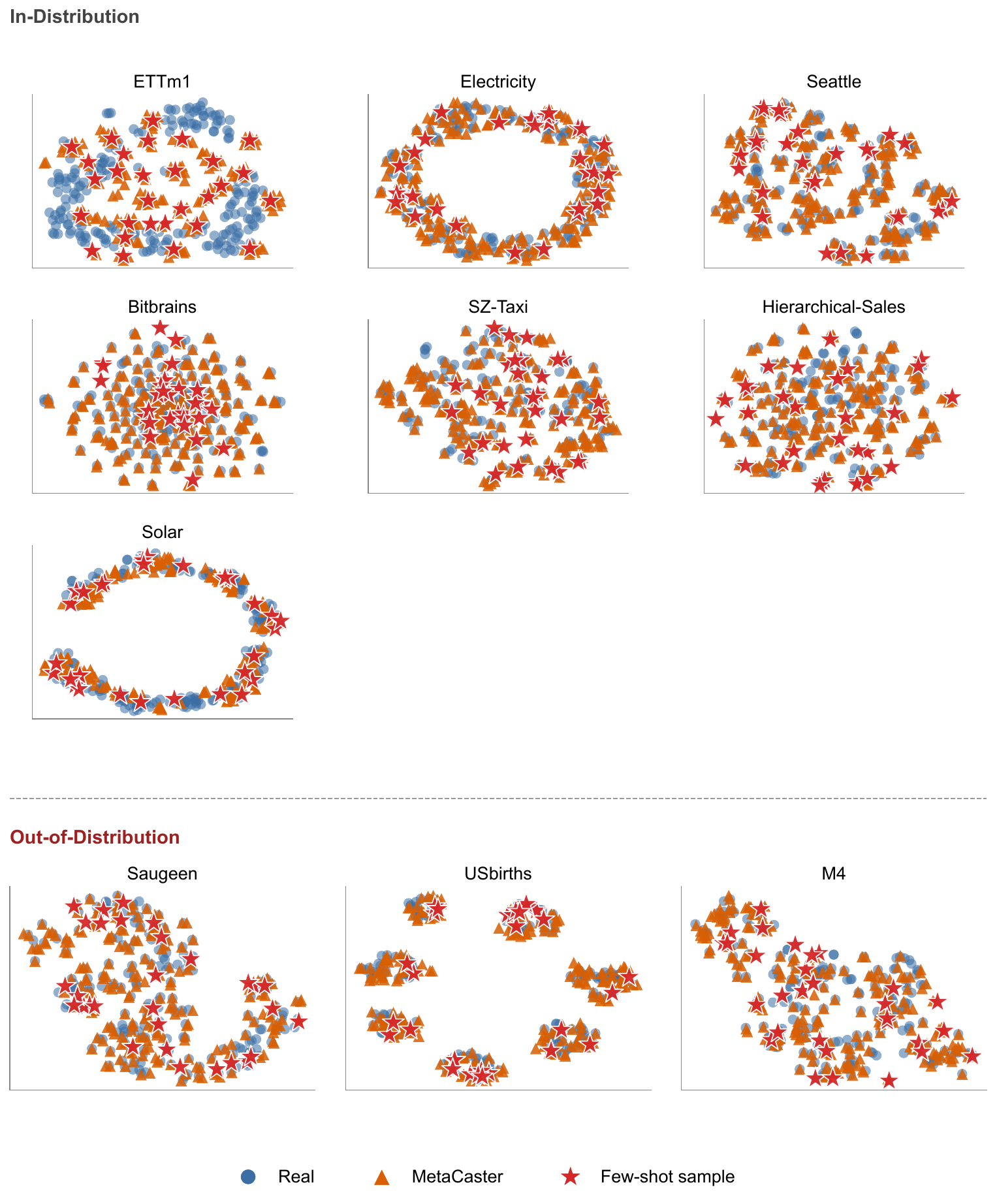}
  \caption{Visualization of data distributions generated by \method on the IND datasets and OOD datasets.}
  \label{fig:tsne-appendix}
\end{figure*}

\paragraph{Visualization of the Generated Data Distributions.}
\label{app:tsne}
Fig.~\ref{fig:tsne-appendix} presents the t-SNE visualization of the distributions of the generated time series by \method. From Fig.~\ref{fig:tsne-appendix}, we observe that the data distributions generated by \method using few-shot examples are mostly consistent with the distributions of the full training datasets in the IND corpus and OOD corpus, suggesting \method's effectiveness in using few-shot examples to generate high-quality time series training datasets.

%% file: Tab/main_results_mae.tex
\begin{table*}[htb]
  \centering
  \footnotesize
  \resizebox{\textwidth}{!}{%
  \begin{tabular}{c l | c | c c c c c | c c c c | c c}
    \toprule
    & \multirow{2}{*}{\textbf{Dataset}}
      & \textbf{Agent}
      & \multicolumn{5}{c|}{\textbf{Generation}}
      & \multicolumn{4}{c|}{\textbf{Augmentation}}
      & \multicolumn{2}{c}{\textbf{Ref}} \\
    & & \method & TimeDP & VerbalTS & T2S & DiffTS & TimeVAE & Repeat & Bootstrap & Jitter & MagWarp & Sample & Full \\
    \midrule
    \multirow{7}{*}{\rotatebox{90}{\textbf{IND ($K=10$)}}}
      & ETTm1       & \best{0.457} & 0.568 & 0.567 & \second{0.531} & 0.582 & 0.620 & 0.650 & 0.637 & 0.636 & 0.930 & 0.625 & 0.400 \\
      & Electricity & \best{0.413} & \second{0.415} & 0.860 & 0.866 & 0.474 & 0.436 & 0.468 & 0.466 & 0.464 & 1.652 & 0.652 & 0.243 \\
      & Seattle     & \second{0.849} & \best{0.844} & 0.989 & 0.989 & 0.971 & 0.948 & 0.995 & 0.962 & 0.986 & 4.395 & 1.127 & 0.663 \\
      & SZTaxi      & \best{0.255} & 0.277 & 0.274 & 0.276 & \second{0.260} & 0.261 & 0.285 & 0.266 & 0.272 & 0.814 & 0.339 & 0.246 \\
      & Sales       & \best{0.661} & 0.795 & 0.771 & 1.027 & 1.510 & 0.830 & 0.945 & 0.766 & \second{0.763} & 1.169 & 0.866 & 0.868 \\
      & Bitbrains   & \second{0.102} & \best{0.085} & 0.114 & 0.110 & 0.135 & 0.149 & 0.172 & 0.150 & 0.156 & 0.408 & 0.208 & 0.148 \\
      & Solar       & \best{0.253} & 0.351 & 0.633 & 0.661 & 0.378 & \second{0.349} & 0.381 & 0.375 & 0.376 & 0.454 & 0.514 & 0.255 \\
    \cmidrule(lr){2-14}
    \multirow{3}{*}{\rotatebox{90}{\textbf{OOD}}}
      & Saugeen     & 0.802 & \best{0.666} & \second{0.675} & 0.743 & 1.003 & 1.042 & 1.108 & 1.044 & 1.053 & 1.184 & 0.816 & 0.548 \\
      & USbirths    & 0.669 & 1.387 & 1.100 & 1.097 & 1.145 & \best{0.605} & 0.665 & \second{0.649} & 0.649 & 2.420 & 0.894 & 0.365 \\
      & M4\textsuperscript{$\dagger$} & \best{1.447} & \second{1.476} & 1.527 & 1.563 & 1.639 & 1.730 & 1.803 & 1.761 & 1.748 & 9.298 & 1.621 & 1.527 \\
    \midrule
    \multirow{7}{*}{\rotatebox{90}{\textbf{IND ($K=30$)}}}
      & ETTm1       & \best{0.424} & 0.570 & 0.572 & \second{0.531} & 0.592 & 0.565 & 0.606 & 0.583 & 0.583 & 0.675 & 0.565 & 0.400 \\
      & Electricity & \best{0.357} & 0.416 & 0.855 & 0.862 & 0.472 & \second{0.385} & 0.411 & 0.398 & 0.401 & 0.824 & 0.616 & 0.243 \\
      & Seattle     & \best{0.740} & \second{0.842} & 0.980 & 0.980 & 0.983 & 1.085 & 0.933 & 0.903 & 0.891 & 3.241 & 0.974 & 0.663 \\
      & SZTaxi      & \second{0.271} & 0.279 & 0.274 & 0.274 & \best{0.263} & 0.300 & 0.315 & 0.295 & 0.290 & 0.999 & 0.309 & 0.246 \\
      & Sales       & \best{0.637} & \second{0.737} & 0.793 & 1.093 & 1.528 & 0.769 & 0.933 & 0.820 & 0.832 & 0.999 & 0.769 & 0.868 \\
      & Bitbrains   & 0.119 & \best{0.087} & 0.116 & \second{0.113} & 0.143 & 0.159 & 0.214 & 0.190 & 0.186 & 0.327 & 0.197 & 0.148 \\
      & Solar       & \best{0.243} & 0.363 & 0.636 & 0.675 & 0.380 & \second{0.340} & 0.356 & 0.353 & 0.349 & 0.405 & 0.478 & 0.255 \\
    \cmidrule(lr){2-14}
    \multirow{3}{*}{\rotatebox{90}{\textbf{OOD}}}
      & Saugeen     & \best{0.647} & \second{0.654} & 0.677 & 0.748 & 1.005 & 0.886 & 0.980 & 0.932 & 0.937 & 0.983 & 0.733 & 0.548 \\
      & USbirths    & 0.570 & 1.218 & 1.104 & 1.090 & 1.132 & \best{0.533} & 0.569 & \second{0.556} & 0.558 & 1.561 & 0.716 & 0.365 \\
      & M4\textsuperscript{$\dagger$} & \best{1.453} & \second{1.499} & 1.547 & 1.588 & 1.632 & 1.845 & 1.768 & 1.763 & 1.780 & 10.04 & 1.718 & 1.527 \\
    \midrule
    \multirow{7}{*}{\rotatebox{90}{\textbf{IND ($K=50$)}}}
      & ETTm1       & \best{0.378} & 0.572 & 0.574 & 0.533 & 0.615 & 0.533 & 0.557 & 0.535 & \second{0.532} & 0.558 & 0.496 & 0.400 \\
      & Electricity & \best{0.328} & 0.409 & 0.855 & 0.885 & 0.465 & \second{0.352} & 0.366 & 0.358 & 0.357 & 0.736 & 0.509 & 0.243 \\
      & Seattle     & \best{0.730} & 0.845 & 0.984 & 0.979 & 1.003 & 1.206 & 0.866 & \second{0.836} & 0.848 & 2.455 & 0.889 & 0.663 \\
      & SZTaxi      & \second{0.265} & 0.275 & 0.274 & 0.269 & \best{0.265} & 0.451 & 0.386 & 0.379 & 0.373 & 0.454 & 0.301 & 0.246 \\
      & Sales       & 0.884 & \second{0.793} & 0.797 & 1.080 & 1.534 & 0.833 & 0.860 & \second{0.793} & \best{0.789} & 0.985 & 0.742 & 0.868 \\
      & Bitbrains   & 0.144 & \best{0.100} & 0.119 & \second{0.113} & 0.136 & 0.186 & 0.317 & 0.281 & 0.292 & 0.381 & 0.180 & 0.148 \\
      & Solar       & \best{0.240} & 0.377 & 0.640 & 0.660 & 0.388 & 0.593 & 0.343 & 0.340 & \second{0.338} & 0.393 & 0.414 & 0.255 \\
    \cmidrule(lr){2-14}
    \multirow{3}{*}{\rotatebox{90}{\textbf{OOD}}}
      & Saugeen     & \best{0.615} & \second{0.642} & 0.678 & 0.747 & 1.007 & 0.803 & 0.970 & 0.940 & 0.939 & 0.950 & 0.716 & 0.548 \\
      & USbirths    & 0.501 & 1.318 & 1.105 & 1.089 & 1.128 & \best{0.459} & 0.476 & 0.464 & \second{0.461} & 1.601 & 0.584 & 0.365 \\
      & M4\textsuperscript{$\dagger$} & \best{1.384} & \second{1.469} & 1.567 & 1.563 & 1.702 & 1.811 & 1.825 & 1.816 & 1.812 & 8.318 & 1.600 & 1.527 \\
    \midrule
    & Wins (of 30) & 19 & 5 & 0 & 0 & 2 & 3 & 0 & 0 & 1 & 0 & -- & -- \\
    \bottomrule
  \end{tabular}%
  }
  \caption{TSF performance comparison on IND and OOD corpora for $K\in\{10, 30, 50\}$ in terms of MAE. Lower MAE is better. \best{Red} (\second{blue}) values indicate the best (second-best) MAE per row. \textsuperscript{$\dagger$}M4 uses instance-normalized MAE to address large distribution shifts.}
  \label{tab:main-mae}
\end{table*}

%% file: Tab/app_topk_per_dataset.tex
\begin{table*}[!t]
  \centering
  \footnotesize
  \resizebox{\textwidth}{!}{%
  \begin{tabular}{l | c | c c c c c | c c c c | c c}
    \toprule
    \textbf{Dataset} & \textbf{Ours} & \multicolumn{5}{c|}{\textbf{Generation Models}} & \multicolumn{4}{c|}{\textbf{Augmentation Methods}} & \multicolumn{2}{c}{\textbf{References}} \\
     & \method & \texttt{TimeDP} & \texttt{VerbalTS} & \texttt{T2S} & \texttt{DiffTS} & \texttt{TimeVAE} & \texttt{Repeat} & \texttt{Bootstrap} & \texttt{Jitter} & \texttt{MagWarp} & $\mathcal{D}_{\text{sup}}^{m}$ & $\mathcal{D}_{\text{tr}}^{m}$ \\
    \midrule
    \multicolumn{13}{l}{\textit{\textbf{Top-1}}} \\
    \midrule
    ETTm1       & \best{0.243} & 0.967 & 0.982 & 0.886 & 0.921 & 0.437 & 0.382 & 0.382 & 0.385 & \second{0.358} & 0.489 & 0.300 \\
    Electricity & 0.148 & 0.209 & 0.947 & 1.017 & 0.237 & 0.128 & 0.125 & 0.125 & \best{0.125} & \second{0.125} & 0.274 & 0.107 \\
    Seattle     & \best{0.949} & 1.271 & 1.658 & 1.674 & 1.288 & 1.678 & 1.097 & \second{1.097} & 1.114 & 1.647 & 1.295 & 0.953 \\
    SZTaxi      & 0.099 & 0.091 & 0.103 & 0.105 & 0.091 & 0.095 & 0.092 & 0.092 & \second{0.091} & \best{0.088} & 0.090 & 0.075 \\
    Sales       & \best{2.176} & 2.265 & \second{2.205} & 2.416 & 2.483 & 2.223 & 2.410 & 2.382 & 2.416 & 2.512 & 2.202 & 2.270 \\
    Bitbrains   & \best{0.056} & \second{0.077} & 0.079 & 0.079 & 0.082 & 0.082 & 0.144 & 0.142 & 0.143 & 0.154 & 0.083 & 0.102 \\
    Solar       & \best{0.135} & 0.173 & 0.453 & 0.503 & \second{0.145} & 0.148 & 0.149 & 0.149 & 0.150 & 0.149 & 0.222 & 0.136 \\
    \cmidrule(lr){1-13}
    Saugeen     & 1.416 & 1.444 & 1.429 & 1.512 & 1.658 & 1.489 & 1.372 & \second{1.371} & 1.380 & \best{1.359} & 1.292 & 1.089 \\
    USbirths    & 0.351 & 1.376 & 1.425 & 1.412 & 1.364 & 0.266 & \second{0.250} & \best{0.249} & 0.255 & 0.502 & 0.487 & 0.106 \\
    M4\textsuperscript{$\dagger$} & \second{1.980} & \best{1.932} & 2.299 & 2.164 & 2.190 & 2.204 & 2.121 & 2.086 & 2.089 & 13.182 & 2.139 & 1.853 \\
    \midrule
    \addlinespace[4pt]
    \multicolumn{13}{l}{\textit{\textbf{Top-3}}} \\
    \midrule
    ETTm1       & \best{0.256} & 0.974 & 0.986 & 0.892 & 0.944 & 0.491 & 0.435 & 0.431 & 0.432 & \second{0.391} & 0.496 & 0.303 \\
    Electricity & 0.158 & 0.219 & 0.974 & 1.026 & 0.271 & \best{0.149} & 0.154 & \second{0.152} & 0.154 & 0.172 & 0.308 & 0.107 \\
    Seattle     & \best{0.985} & 1.295 & 1.698 & 1.696 & 1.363 & 1.697 & 1.177 & \second{1.151} & 1.192 & 2.052 & 1.351 & 0.966 \\
    SZTaxi      & 0.101 & 0.095 & 0.105 & 0.106 & \best{0.092} & 0.098 & 0.097 & \second{0.094} & 0.095 & 0.115 & 0.106 & 0.077 \\
    Sales       & \best{2.201} & 2.295 & \second{2.225} & 2.464 & 2.634 & 2.248 & 2.504 & 2.437 & 2.486 & 2.597 & 2.221 & 2.324 \\
    Bitbrains   & \best{0.068} & \second{0.077} & 0.079 & 0.079 & 0.085 & 0.083 & 0.178 & 0.171 & 0.159 & 0.173 & 0.085 & 0.104 \\
    Solar       & \best{0.137} & 0.181 & 0.470 & 0.509 & \second{0.163} & 0.165 & 0.172 & 0.167 & 0.172 & 0.183 & 0.227 & 0.137 \\
    \cmidrule(lr){1-13}
    Saugeen     & 1.422 & 1.450 & 1.435 & 1.535 & 1.712 & 1.572 & \best{1.395} & \second{1.396} & 1.404 & 1.572 & 1.306 & 1.096 \\
    USbirths    & 0.369 & 1.452 & 1.457 & 1.417 & 1.396 & \best{0.326} & 0.336 & \second{0.332} & 0.334 & 0.695 & 0.513 & 0.123 \\
    M4\textsuperscript{$\dagger$} & \second{2.006} & \best{1.974} & 2.311 & 2.168 & 2.278 & 2.506 & 2.219 & 2.208 & 2.280 & 34.704 & 2.186 & 1.974 \\
    \midrule
    \addlinespace[4pt]
    \multicolumn{13}{l}{\textit{\textbf{Top-5}}} \\
    \midrule
    ETTm1       & \best{0.282} & 0.979 & 0.989 & 0.894 & 0.970 & 0.539 & 0.493 & 0.467 & 0.469 & \second{0.466} & 0.499 & 0.304 \\
    Electricity & \best{0.183} & 0.229 & 0.998 & 1.040 & 0.293 & \second{0.192} & 0.200 & 0.194 & 0.196 & 0.263 & 0.337 & 0.107 \\
    Seattle     & \best{1.046} & 1.314 & 1.730 & 1.716 & 1.403 & 1.704 & 1.280 & \second{1.249} & 1.268 & 5.347 & 1.403 & 0.971 \\
    SZTaxi      & 0.102 & \second{0.098} & 0.106 & 0.107 & \best{0.095} & 0.102 & 0.102 & 0.098 & 0.098 & 0.185 & 0.112 & 0.079 \\
    Sales       & \best{2.213} & 2.311 & \second{2.234} & 2.563 & 2.787 & 2.273 & 2.537 & 2.481 & 2.518 & 2.721 & 2.241 & 2.381 \\
    Bitbrains   & \best{0.074} & \second{0.077} & 0.079 & 0.082 & 0.090 & 0.085 & 0.195 & 0.188 & 0.181 & 0.211 & 0.089 & 0.118 \\
    Solar       & \best{0.141} & 0.185 & 0.477 & 0.517 & 0.187 & \second{0.183} & 0.190 & 0.186 & 0.190 & 0.226 & 0.234 & 0.138 \\
    \cmidrule(lr){1-13}
    Saugeen     & \best{1.425} & 1.455 & \second{1.438} & 1.543 & 1.767 & 1.595 & 1.623 & 1.598 & 1.619 & 1.839 & 1.320 & 1.104 \\
    USbirths    & 0.394 & 1.478 & 1.467 & 1.429 & 1.429 & \best{0.358} & 0.372 & \second{0.359} & 0.368 & 1.051 & 0.537 & 0.133 \\
    M4\textsuperscript{$\dagger$} & \second{2.013} & \best{2.000} & 2.319 & 2.181 & 2.331 & 2.667 & 2.273 & 2.267 & 2.358 & 52.066 & 2.253 & 2.021 \\
    \bottomrule
  \end{tabular}%
  }
  \caption{TSF performance on test corpora for top selected forecasters by the compared methods in terms of MSE. Lower MSE is better. \best{Red} (\second{blue}) values indicate the best (second-best) MSE per row. \textsuperscript{$\dagger$}M4 uses instance-normalized MSE to address large distribution shifts.}
  \label{tab:topk-per-dataset}
\end{table*}